\documentclass[conference]{IEEEtran}
\IEEEoverridecommandlockouts
\usepackage{cite}
\usepackage{amsmath,amssymb,amsfonts}
\usepackage{algorithmic}
\usepackage{graphicx}
\usepackage{textcomp}
\usepackage{xcolor}
\usepackage{hyperref}
\usepackage{balance}
\usepackage[ruled,vlined,linesnumbered]{algorithm2e}
\def\BibTeX{{\rm B\kern-.05em{\sc i\kern-.025em b}\kern-.08em
    T\kern-.1667em\lower.7ex\hbox{E}\kern-.125emX}}
\usepackage{subfigure}
\begin{document}

\title{ORIS: Online Active Learning Using Reinforcement Learning-based Inclusive Sampling for Robust Streaming Analytics System\\
}

\author{\IEEEauthorblockN{Rahul Pandey\textsuperscript{*}\thanks{\textsuperscript{*}The author is currently with Amazon.}}
\IEEEauthorblockA{
\textit{George Mason University}\\
Fairfax VA, USA \\
rpandey4@gmu.edu}
\and
\IEEEauthorblockN{Ziwei Zhu}
\IEEEauthorblockA{
\textit{George Mason University}\\
Fairfax VA, USA \\
zzhu20@gmu.edu}
\and
\IEEEauthorblockN{Hemant Purohit}
\IEEEauthorblockA{
\textit{George Mason University}\\
Fairfax VA, USA \\
hpurohit@gmu.edu}
}
\maketitle

\begin{abstract}
Effective labeled 
data collection 
plays a critical role in developing 
and fine-tuning 
robust streaming analytics systems. 
However, 
continuously labeling documents 
to filter relevant information poses significant challenges like limited labeling budget or lack of high-quality labels. 
There is a need for efficient human-in-the-loop machine learning (HITL-ML) design to 
improve 
streaming analytics systems. One particular HITL-ML approach is online active learning, which involves iteratively selecting a small set of the most informative documents for labeling to enhance the 
ML model performance. 
The performance of 
such 
algorithms can get affected due to human errors in labeling. 
To address these challenges, we propose \emph{\textbf{ORIS}}, a method to perform \emph{\textbf{O}}nline active learning using  \emph{\textbf{R}}einforcement learning-based \emph{\textbf{I}}nclusive \emph{\textbf{S}}ampling 
of documents for labeling. 
ORIS aims to create a novel Deep Q-Network-based strategy to sample incoming documents that minimize human errors in labeling and enhance the ML model performance. We evaluate the 
ORIS method on emotion recognition tasks, and it outperforms traditional baselines in terms of both human labeling performance and the ML model performance. The code for this research 
is available at \href{https://github.com/rpandey4/oris}{https://github.com/rpandey4/oris}.
\end{abstract}


\section{Introduction}
\label{sec:intro}
\begin{figure*}
	\centering
		\includegraphics[scale=0.125]{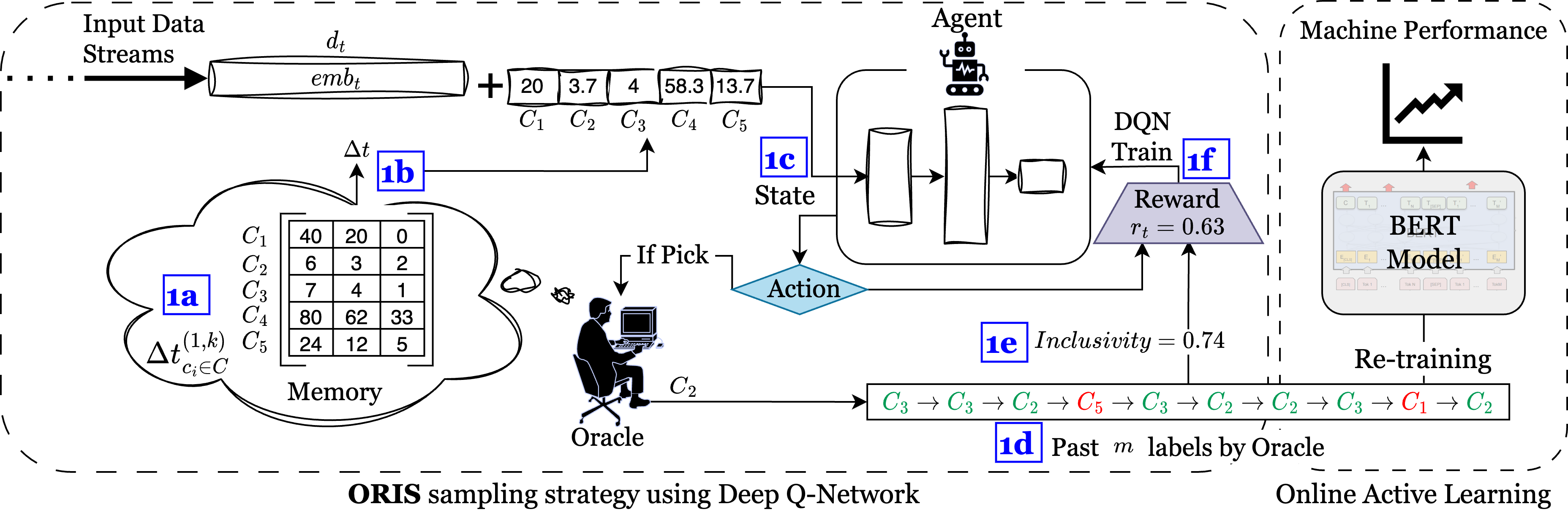}
	\caption{Overview of the proposed ORIS architecture. The method aims to sample inclusive documents from streams using a DQN-based agent. Components 1a, 1b, \& 1c are described in Sec.~\ref{subsec:oris_m_state}, while Components 1d, 1e, \& 1f are detailed in Sec.~\ref{subsec:oris_m_reward}.
 }
	\label{fig:oris}
 \vspace{-2mm}
\end{figure*}
The exponential growth of online data for various application domains, such as 
journalism, public health, and crisis management, has presented new challenges in effectively filtering and processing high-volume, high-velocity data streams. These data streams are often characterized by their noisy, sparse, and redundant nature, making it difficult for human annotators to keep pace with the sheer velocity and volume of data~\cite{castillo2016big}. 
Further, 
purely automated systems for streaming analytics face limitations in accurately filtering data and fail to adapt to the dynamic nature of the data. 

To address these challenges, human-in-the-loop machine learning (HITL-ML) methods like online active learning have emerged 
that combines human labeling and automated classification to achieve accurate and efficient data filtering~\cite{gama_survey_2014,ren_survey_2021}. Online active learning methods 
selectively request labels for informative documents from a human (oracle), reducing the overall labeling cost while maintaining ML model accuracy. This approach has shown success in various tasks, such as object detection, image/video/text classification, and machine translation systems~\cite{imran_engineering_2014, lofi_design_2014}.


However, the accuracy and reliability of labeling can get affected by various human factors during the labeling process. 
For instance, 
prior research~\cite{pandey2019modeling, pandey_modeling_2022} has shown the presence of serial ordering-induced human errors like Mistakes and Slips~\cite{reason_human_2000} in the case of labeling task. 
Mistakes result from the absence of a correct cognitive representation of a concept, while slips 
occur despite acquiring the correct cognitive representation of a concept due to memory decay over time. 
To mitigate the serial ordering-induced slip error, researchers have proposed a heuristic-driven approach to sample the documents that can reduce the error in labeling and increase the annotator reliability, which in turn increases the performance of the ML model~\cite{pandey2019modeling, pandey_modeling_2022}. However, there are certain challenges in using heuristic-driven approaches, like inflexibility to new tasks, increased bias, and limited scalability. Thus, a dynamic data-driven technique like reinforcement learning emerges as an indispensable and urgently required solution to address the challenge of minimizing the slips type of human error for the labeling tasks on data streams. 

We propose \emph{\textbf{ORIS}}, a method to perform \emph{\textbf{O}}nline active learning using  \emph{\textbf{R}}einforcement learning-based \emph{\textbf{I}}nclusive \emph{\textbf{S}}ampling. ORIS aims to minimize human errors in labeling and enhancing the performance of the ML model. We use a novel Deep Q-Network~\cite{mnih_human-level_2015} based strategy for reinforcement learning to sample incoming documents in a data stream efficiently that leads to robust active learning. We introduce a novel state representation and reward function, which learns the policy of inclusive sampling-based active learning for streaming data. The learned policy helps in the efficient collection of high-quality streaming data that reduces human error in labeling and improves the ML model quality. 

Our contributions in this paper are: 1) We formulate the problem of human error in a streaming analytics 
framework. In this framework, a oracle annotator mimicking memory decay behavior is used for labeling that can pass erroneous labels. 2) We propose a novel online active learning method, ORIS, which is capable of sampling documents that are inclusive and 
less prone to human error (see Fig.~\ref{fig:oris}). 3) With extensive experimentation and evaluation to compare the proposed ORIS method with the baselines on emotion recognition tasks, we achieve up to $38.3\%$ human \& $55.7\%$ machine performance improvement on \textit{Twitter (now X.com)}, and up to $44.2\%$ human \& $70.1\%$ machine performance improvement on \textit{Reddit}.

\section{Related Work}
\label{sec:related}
\subsection{Online Active Learning}
\vspace{-2mm}
\label{rel:oal}
Traditional batch-based Active Learning (AL) assumes the existence of a pool of unlabeled data from which to select the most informative documents for labeling~\cite{rudovic_multi-modal_2019, yuan_cold-start_2020}. However, in streaming analytics systems, data comes in real-time streams. Hence, the decision to pick or discard a document must be made in real-time, making batch-based AL impractical. For example, in the case of processing sequential online real-world data such as social media streams, gathering the true label is both costly and time-consuming. Hence, prior work has proposed an online active learning that samples data coming in streams for efficient model training~\cite{cheng_feedback-driven_2013,hao2018online, zliobaite_active_2014}.
\vspace{-2mm}
\subsection{Deep Reinforcement Learning for Active Learning}
\vspace{-2mm}
Traditional active learning depends on heuristic-driven approaches for coming up with sampling strategies. However, researchers have explored the use of non-heuristic methods in active learning, such as reinforcement learning, to automate the design of deep learning models and active learning query strategies~\cite{ren_survey_2021}. 
For example, Deep Reinforcement Active Learning (DRAL), applies the idea of reinforcement learning to dynamically adjust the acquisition function for specific tasks such as named entity recognition~\cite{fang_learning_2017}, person re-identification~\cite{liu_deep_2019}, image segmentation~\cite{casanova_reinforced_2019}, and multimodal classification~\cite{rudovic_multi-modal_2019}. The DRAL framework selects sequential documents from a gallery pool or streaming input during the active learning process, obtaining manual labels with binary/multi-class feedback. The rewards and the oracle feedback are used to adjust the agent's queries, ensuring the selection of high-quality query samples. 
This approach enables more flexible and efficient active learning processes, improving the accuracy and reliability of labeling while reducing the workload on human annotators.

\vspace{-2mm}
\subsection{Human Factors in Data Labeling}
\vspace{-2mm}
Prior works have utilized psychology literature to improve labeling. For example, \cite{park_ai-based_2019} uses social strategies of interactions from psychology literature to improve crowdsourcing participation. By augmenting questions in 
visual question answering task using the social strategies, they have increased the crowd workers' participation and informative responses. Researchers have also found a direct correlation between the monetization of the crowdsourcing tasks with the labeling speed \cite{vaughan_making_2018}. 
Most recent research studied the memory decay behavior from psychology in the case of labeling quality,
explained in detail in Section~\ref{sec:pre_mem}~\cite{pandey2019modeling, pandey_modeling_2022}.

\section{Preliminaries}
\label{sec:prelim}

\subsection{Online Active Learning}
\begin{figure}
	\centering
		\includegraphics[scale=0.12]{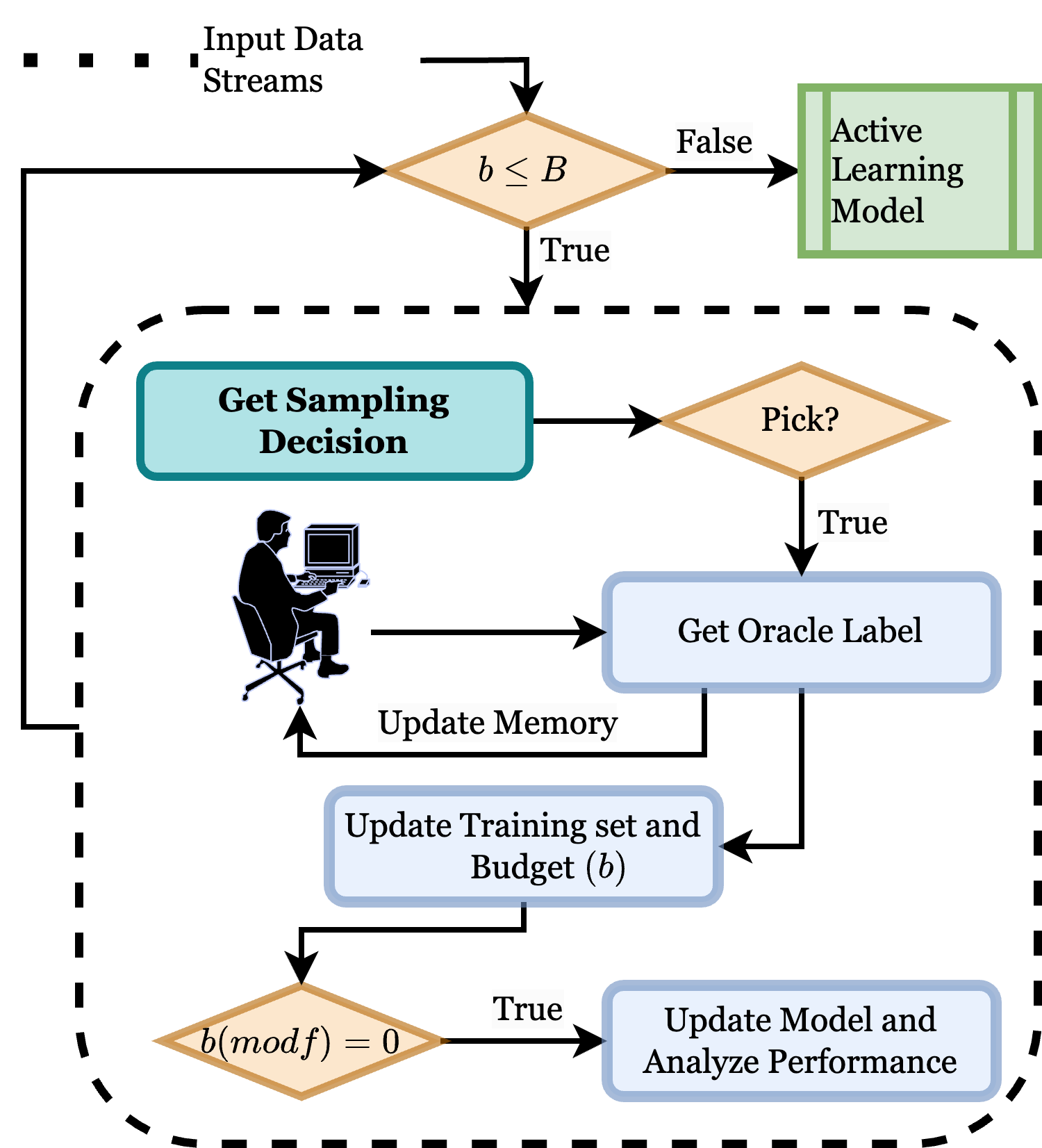}
	\caption{Online Active Learning System Flow.
 }
	\label{algo:oal}
 \vspace{-6mm}
\end{figure}

\vspace{-2mm}
As discussed in Section~\ref{rel:oal}, online active learning is a promising approach for minimizing the labeling cost while improving the ML model's performance. 
In online active learning, 
 the ML model 
retrains with the new labeled documents added to the training set. This process continues until a stopping criterion is met, such as reaching a certain performance threshold or exhausting the labeling budget. Fig.~\ref{algo:oal} shows the generic flow of online active learning. It assumes the data is coming in streams in real-time. There is a fixed set of test documents to analyze the performance of the active learning model ($ALM$), and the update frequency $f$ decides when to update the $ALM$ model (using the condition $b(modf) = 0$). The whole system runs till the maximum budget of $B$ to receive the oracle's feedback. Given the current document, we get the decision to sample document for labeling from an agent. This agent can be heuristic-driven or data-driven. If the agent's decision is to pick, then we request oracle to provide the label. We keep an update of the past selected labels in the oracle memory for mimicking the oracle behavior as discussed in the next Section~\ref{sec:pre_mem}. Moreover, we update the training set by including the current document-label pair, which will be used for retraining the $ALM$. At every update frequency $f$ of the budget exhausted, we re-train the $ALM$ model with the training set. Moreover, we analyze the performance on the independent test documents.

\subsection{Human Memory and Labeling Error}
\label{sec:pre_mem}
\vspace{-2mm}
The memory decay behavior of humans is widely studied in psychology literature, which shows that memory retention decreases exponentially with time. Prior research~\cite{reason_human_2000} provides a human error taxonomy that distinguishes between two classes of errors: mistakes and slips as discussed in the Introduction. 

Prior research showed that a similar taxonomy exists in the crowdsourcing domain~\cite{pandey2019modeling, pandey_modeling_2022}. 
They studied the memory decay behavior of humans in the context of learning and acquiring new knowledge, which results in the serial ordering-induced mistakes and slips~\cite{brown1958some,ebbinghaus1913memory}. Psychologists have used an exponential function in the past to model the memory decay of humans~\cite{anderson1991reflections,loftus_evaluating_1985}. In the context of labeling, mathematically, the probability of an oracle making an error in labeling a particular class label $c$ is defined using an exponential function over time last seen ($\Delta t_c$). Prior work used a parameterized sigmoid function~\cite{pandey2019modeling,pandey_modeling_2022} to compute the error probability score as defined in Eq~\ref{eqn:err_sig}: 
\begin{equation}
\small
    \text{error\_probability\_score}(c) = \frac{1}{1+e^{-\alpha \Delta t_c + \beta}}
    \label{eqn:err_sig}
\end{equation}
where $\alpha$ and $\beta$ are hyperparameters representing different decay intensities of humans. Finally, we extend a new parameterized exponential decay function in an oracle to further strengthen the efficacy of the proposed method.
We use an exponential function to compute the error probability score 
as defined in Eq~\ref{eqn:err_exp}: 
\begin{equation}
\small
    \text{error\_probability\_score}(c) = max(1, e^{\alpha \Delta t_c + \beta})
    \label{eqn:err_exp}
\end{equation}



\section{Approach}
In this section, we introduce the problem of streaming analytics systems for text document classification in the presence of a error-prone oracle annotator. We then explain the design of our ORIS method. Note that ORIS can be extended to other modalities, such as images \& videos, and other streaming analytics tasks. 
\subsection{Problem Formulation}
\vspace{-1mm}
Given a stream of documents ($d_1, d_2, ..., d_t, ...$) arriving in real-time, our objective is to sample documents for labeling by the oracle to improve the ML model ($ALM$). The maximum budget to pick documents to train is defined by the hyperparameter $B$. Whenever a document is picked, we request a label from the oracle. The oracle induces the serial order-induced slip error behavior in providing the annotation as described in Section~\ref{sec:pre_mem}. Once we receive the label, we store it in the training set $S$ for re-training the ML model. We also keep track of the prior labeling memory $M$ to correctly imitate the real-world system where oracle posses memory decay behavior. We (re-)train a BERT-based ML model ($ALM$) with the training set $S$, and the frequency of its retraining is defined by hyperparameter $f$ $(f < B)$. Moreover, after each retraining with frequency $f$, we observe the ML model's ($ALM$) performance (machine performance) on an independent test set ($T$).  Furthermore, we observe the human performance of the oracle's behavior in providing the correct or erroneous labels.
We report both performance metrics at every frequency interval $f$.

\begin{algorithm*}
\footnotesize
\SetAlgoLined
\caption{Deep Q-Network Training for ORIS}
\label{algo:oris_train}
\SetKwInOut{Input}{Input}\SetKwInOut{Output}{Output}
\Input{Dataset $D$, Replay buffer $\mathcal{R}$, Q-network with weights $\theta$, target network with weights $\theta^{-}$, exploration rate $\epsilon$, discount factor $\gamma$, budget $B$, maximum number of episodes $E_{max}$}
\Output{Optimal Q-values $Q^{*}$}
Initialize $\mathcal{R}$ with capacity $N$\;
Initialize $\theta$ and $\theta^{-}$ with random weights\;
\For{$i = 1$ to $E_{max}$}{
    $D^{i} \gets Shuffle(D)$;
    $t \gets 1$;
    $b \gets 1$;
    $M_t \gets InitializeOracleMemory()$;
    $\Delta t_t \gets InitializeTimeLastSeen()$;
    
    \While{$b \leq B$}{
    $d_t \gets GetDocument(D^i, t)$;
    $emb_{t} \gets GetEmbedding(d_t)$;
    $s_{t} \gets Concat(emb_t, \Delta t_t)$;
    $a_t \leftarrow {\mathrm{argmax}}\ Q(s_{t}, a; \theta) $
    
    \If{$a_t == 1$}{
        $b \gets b + 1$;
        $c_t \gets GetOracleLabel(d_t)$;
        $M_t \gets UpdateOracleMemory(M_t, c_t)$;
        $\Delta t_{t+1} \gets UpdateTimeLastSeen(\Delta t_t, c_t)$
    }
    $r_{t+1} \gets ComputeReward(a_t, M_t);$
    $emb_{t+1} \gets GetEmbedding(D^{i}_{t+1})$;
    $s_{t+1} \gets Concat(emb_{t+1}, \Delta t_{t+1})$;
    
    Store transition $(s_{t}, a_{t}, r_{t+1}, s_{t+1})$ in $\mathcal{R}$\;
    Sample a minibatch of transitions 
    from $\mathcal{R}$\;
    Set $y_{t} = r_{t+1} + \gamma \underset{a}{\mathrm{max}}\ Q(s_{t+1}, a; \theta^{-})$\;
    Update $\theta$ by minimizing the loss $L(\theta) = SmoothL1Loss(y_{t}, Q(s_{t}, a_{t}; \theta))$\;
    Update the target network weights: $\theta^{-} \leftarrow \tau * \theta + (1 - \tau) * \theta^{-}$\;
    $t \gets t + 1$;
    }
}
\end{algorithm*}








    


\subsection{Overview of ORIS}
\vspace{-1mm}
In this section, we describe the proposed ORIS method used for inclusive sampling. Unlike relying on a heuristic-based approach, ORIS uses reinforcement learning to develop the sampling strategy as used in recent prior work on deep reinforcement active learning (DRAL)~\cite{casanova_reinforced_2019, fang_learning_2017, liu_deep_2019, rudovic_multi-modal_2019}. However, the objective of ORIS is different from the prior DRAL-based approaches. Instead of only focusing on the machine performance of the ML model, the proposed ORIS method tackles both human and machine performance to create a robust active learning sampling strategy. 

Fig.~\ref{fig:oris} shows the overall architecture of ORIS. It comprises two components: ORIS sampling using Deep Q-Network and Online Active Learning. To create an optimal sampling strategy, we formulate the problem of sampling the documents coming in streams as a Markov Decision Process (MDP)~\cite{bellman1957markovian}. Given the current document $d_t$ at time $t$, we represent using document embedding. Additionally, we model the current memory of the oracle to include it in the state variable along with the document embedding. We initialize Deep Q-Network (DQN)~\cite{mnih_human-level_2015}, which consists of a feed-forward neural network as a Q-network, and acts as a reinforcement learning agent that takes the state as input and outputs the action.
The DQN agent takes action on the incoming documents $d_t$ till the budget $B$ to pick the documents is exhausted. We store each experience containing the state, action, reward, and next state in the replay buffer $\mathcal{R}$. During training, we utilize these experiences to minimize the mean squared error between the predicted Q-value and the target Q-value. We train the DQN for several episodes with different online real-time streaming
till maximum episode $E_{max}$. Once the DQN is trained, we use it for online active learning's  decision-making. 
\subsection{ORIS Modeling}
\vspace{-1mm}
In this section, we explain the different components for modeling the proposed ORIS method as an MDP problem. First, we define the formation of state, action, and reward, followed by detailed information on forming the DQN architecture, the training, and the inference. 
\subsubsection{State}
\label{subsec:oris_m_state}
The state variables consist of the representation of the environment, which is the input to the DQN to get the optimal action. Given the incoming streaming input text $d_t$, we compute its embeddings $emb_t$ as a part of the state variable. We use the pre-trained word embeddings to compute the document embedding representation. 
Consider the document $d_t$ as a set of words $[w_1, w_1, ..., w_n]$ of length $n$, we compute embeddings at time $t$ as $emb_t = \frac{\sum_{w \, \epsilon \, d_t} emb^{w}}{n}$ where $emb^{w}$ is the pre-trained word embeddings of the word $w$. 

Moreover, we keep track of the prior class labels from the oracle as an additional input of the state variable. This input helps the agent to understand the oracle memory and its effect on the label quality to make an inclusive decision to pick or discard the current document. Given the current step $t$, we compute the latest time last seen for class $c_i \in C$ by the oracle as $\Delta t^{1}_{c_i} = t - j_1$, where $j_1$ is the latest step at which $c_{j_1} == c_i$. However, there exists a possibility where an oracle can forget a class and make errors in labeling, and hence the class label used to compute the time last seen may not be valid. Hence, to ensure the reliability of the time last seen value computation, we keep track of the most recent $k$ time last seen values for each class $c_i \in C$. Fig.~\ref{fig:oris} (1a) shows the example of keeping the time last seen by the oracle of the last $k=3$ occurrences of a five-class labeling task. 
The updated time last seen values become $\Delta t_{c_i} = \frac{\sum_{j=1}^{k}\Delta t^{j}_{c_i}}{k}$ where $j$ denotes the $j^{th}$ latest step oracle has seen the class $c_i$ (\textit{c.f.} Fig.~\ref{fig:oris} (1b)). 

Finally, the state variable $s_t$ is the concatenation of current input text embedding $emb_t$ and the $k$-averaged time last seen for all the classes as shown in Fig.~\ref{fig:oris} (1c): 
\begin{equation}
\small
\begin{aligned}
    \Delta t_t = Concat(\Delta t_{c_i}) \; \; \forall \; c_i \; in \; C \\
    s_t = Concat(emb_t, \Delta t_t) 
\end{aligned}
\label{eqn:state}
\end{equation}

The length of the state $s_t$ is equal to $len(emb_t)$ (length of embedding) $+$ $ len(C)$ (total classes). 
Fig.~\ref{fig:oris} illustrates an example of state variable computation. The incoming document is $d_t$ with document embedding $emb_t$. There are five classes to label. The table in the figure represents the time last seen $\Delta t_{c_i}$ corresponding to the past $k=3$ last seen for each class $c_i$. The $\Delta t_t$ is calculated as the $k$-averaged time last seen for every class and concatenated with the embedding $emb_t$ to make the state variable representation $s_t$.  

\subsubsection{Action} 
Since the ORIS agent aims to sample the input documents, the action has two values: pick $(1)$ if labeling the document and discard $(0)$ if not. 
If 
picked, the oracle 
provides the label, which is used to retrain the ML model (ref. Fig.~\ref{algo:oal}). 
\subsubsection{Reward}
\label{subsec:oris_m_reward}
The reward incentivizes the optimal behavior of the DQN model. Since the objective of ORIS is not to forget any classes by the oracle, we reward the agent if the selected documents are equally diversified along with their class labels. Note that the oracle can make no error in labeling if all the sampled documents belong to the same class. However, that causes risk to both the oracle and ML model. The ML model will not learn well for all classes if all the picked documents for training belong to the same class. Moreover, the oracle will have the highest risk of providing errors if the selected documents belong to any other class besides the frequent one. 

Hence, the proposed reward function promotes diversity and inclusivity of all classes of recently sampled documents. We note that the diversity in sampling not only reduces the chance of the oracle making fewer serial ordering-induced slips errors but also improves the 
performance of the ML model, especially for the infrequent class in the imbalanced data distribution, which is the case for many real-time streaming tasks. To achieve this, we introduce a new intermediate metric $Inclusivity$, which measures the Shannon entropy~\cite{shannon_mathematical_2001} of the annotated classes of past $m$ picked documents in the Memory $M$ as shown in Eq~\ref{eqn:reward}. 
\begin{equation}
\small
\begin{aligned}
    Inclusivity(M) = -\sum_{i=1}^{C} p^M(c_i) \log_{2} p^M(c_i) \\
    ComputeReward(a, M) = \left\{\begin{matrix}
    \rho *e^{\delta (Inclusivity(M) - 1)))} & a = 1\\ 
    \lambda & a = 0
    \end{matrix}\right.
\end{aligned}
\label{eqn:reward}
\end{equation}

Moreover, to make the overall reward promote higher inclusivity and penalize marginal or low inclusivity scores, we use a parameterized exponential function as shown in Eq~\ref{eqn:reward}. The parameter $\delta$ helps to deactivate the marginal inclusivity (entropy) score and amplify the high entropy score. The parameter $\rho$ helps to amplify the reward values to $[0, \rho)$. 
Fig.~\ref{fig:oris} (1d, 1e, \& 1f) 
shows the example of computing reward score given the past $m$ labels. 
\subsection{ORIS Training and Inference}
\vspace{-1mm}
Once we model the ORIS as an MDP problem, we train it using the DQN training strategy~\cite{mnih_human-level_2015}. Our goal is to learn an optimal policy such that the oracle makes fewer errors, and the ML model is trained with diverse error-free documents. 
Algorithm~\ref{algo:oris_train} describes the training procedure of our proposed ORIS method. It takes a set of documents D as input, which we use to create different streaming scenarios. Moreover, similar to the original approach of DQN~\cite{mnih_human-level_2015}, we use a replay buffer $\mathcal{R}$ to store the state-action transition $(s_t, a_t, r_{t+1}, s_{t+1})$. We also initialize two Q-networks: source and target, each with weights $\theta$ and $\theta^{-}$, respectively (Line 2). It consists of three dense layers with ReLU activation as 
in Eq~\ref{eqn:dqn-network}. 
\begin{equation}
\small
\begin{aligned}
h_{l1} = ReLU(Dense(s_t))
\\
h_{l2} = ReLU(Dense(h_{l1}))
\\ 
\text{Q-Value} = Dense(h_{l2})
\end{aligned}
\label{eqn:dqn-network}
\end{equation}
An episode is calculated once the agent has picked the documents equivalent to the budget of $B$. 
We train the DQN for several episodes until  $E_{max}$. At each episode, we shuffle the dataset $D$ to explore different real-time streaming scenarios (Line 4).
We keep track of all picked documents and its label in memory $M_t$. We use $M_t$ to compute the time last seen $\Delta t$ for state representation (\textit{c.f.} Eq~\ref{eqn:state}) and to compute the reward (\textit{c.f.} Eq~\ref{eqn:reward}). Once we create real-time streaming data $D^{i}$ for episode $i$, we extract each document $d_t$ and compute the state representation $s_t$ for the agent (Line 6). To make the DQN explore different actions, we keep an exploration rate $\epsilon$ and exponentially decrease it over time. We use this exploration rate as a probability to choose random action as opposed to greedy action based on the Q-value as shown in Line 6.
If the action $a_t = 1$ (pick), we fetch the oracle label for the current input $d_t$ (Line 8). During training, we do not 
consider 
error-prone oracle, as we want to keep the training unbiased from a specific memory decay behavior. 
Next, we update oracle memory in $M_t$, and we update the time last seen ($\Delta t_{t+1}$) based on the current label $c_t$ (Line 8). Note that we only update the $\Delta t_{t+1}$ if the action is to pick; else, we keep it the same as the previous value. We also update the budget-exhausted counter $b$ for the current episode $i$. 
Based on the action $a_t$ and the current memory $M_t$, we compute the reward using Eq~\ref{eqn:reward} (Line 10). 
Next, we compute the succeeding state representation to store transition in the replay buffer $\mathcal{R}$ (Line 10-11). To train DQN, we sample a minibatch of transitions from $\mathcal{R}$ and compute the predicted Q-value (Line 12-13). We update source Q-network weights $\theta$ by minimizing the smooth L1 loss between the predicted Q-value and the source Q-value (Line 14). We also soft update the target Q-network weights $\theta^{-}$ with the weights of source Q-network $\theta$ using the factor $\tau$ at every step (Line 15).
During inference, we use the trained source Q-network with weights $\theta$ as \textit{`Get Sampling Decision'} function in Fig.~\ref{algo:oal} to perform the robust online active learning sampling.



\begin{table}  
\centering
\small
\caption{Dataset for experiments of RL training and online active learning for both Reddit and Twitter data.
}
\label{tab:dataset}
\resizebox{\columnwidth}{!}{\begin{tabular}{|l|c|c|c|c|c|c|}
\hline
\textbf{Type}& \textbf{Total} & \textbf{sadness} & \textbf{joy} & \textbf{surprise} & \textbf{anger} & \textbf{fear} \\ \hline \hline
\multicolumn{7}{|l|}{\textbf{Reinforcement Learning Training}} \\ \hline
\textbf{rl-train} & 362210         & 115154     & 133737     & 13982      & 54368      & 44969      \\ \hline \hline
\multicolumn{7}{|l|}{\textbf{Twitter Online Active Learning}} \\ \hline
\textbf{train} & 14696          & 4666       & 5362       & 572        & 2159       & 1937       \\ \hline
\textbf{test} & 1841           & 581        & 695        & 66         & 275        & 224        \\ \hline \hline
\multicolumn{7}{|l|}{\textbf{Reddit Online Active Learning}} \\ \hline
\textbf{train}  & 3845           & 817        & 853        & 720        & 1025       & 430        \\ \hline
\textbf{test}  & 478           & 102        & 93         & 87         & 131        & 65         \\ 
\hline
\end{tabular}}
\vspace{-4mm}
\end{table}
\section{Experiments}
\label{sec:exp}
\begin{figure*}[ht]
\centering
\setlength{\tabcolsep}{2pt}
\renewcommand{\subfigcapmargin}{0pt}
\renewcommand{\subfigbottomskip}{0pt}
\renewcommand{\subfigtopskip}{0pt}
\subfigure{\includegraphics[width=0.23\textwidth]{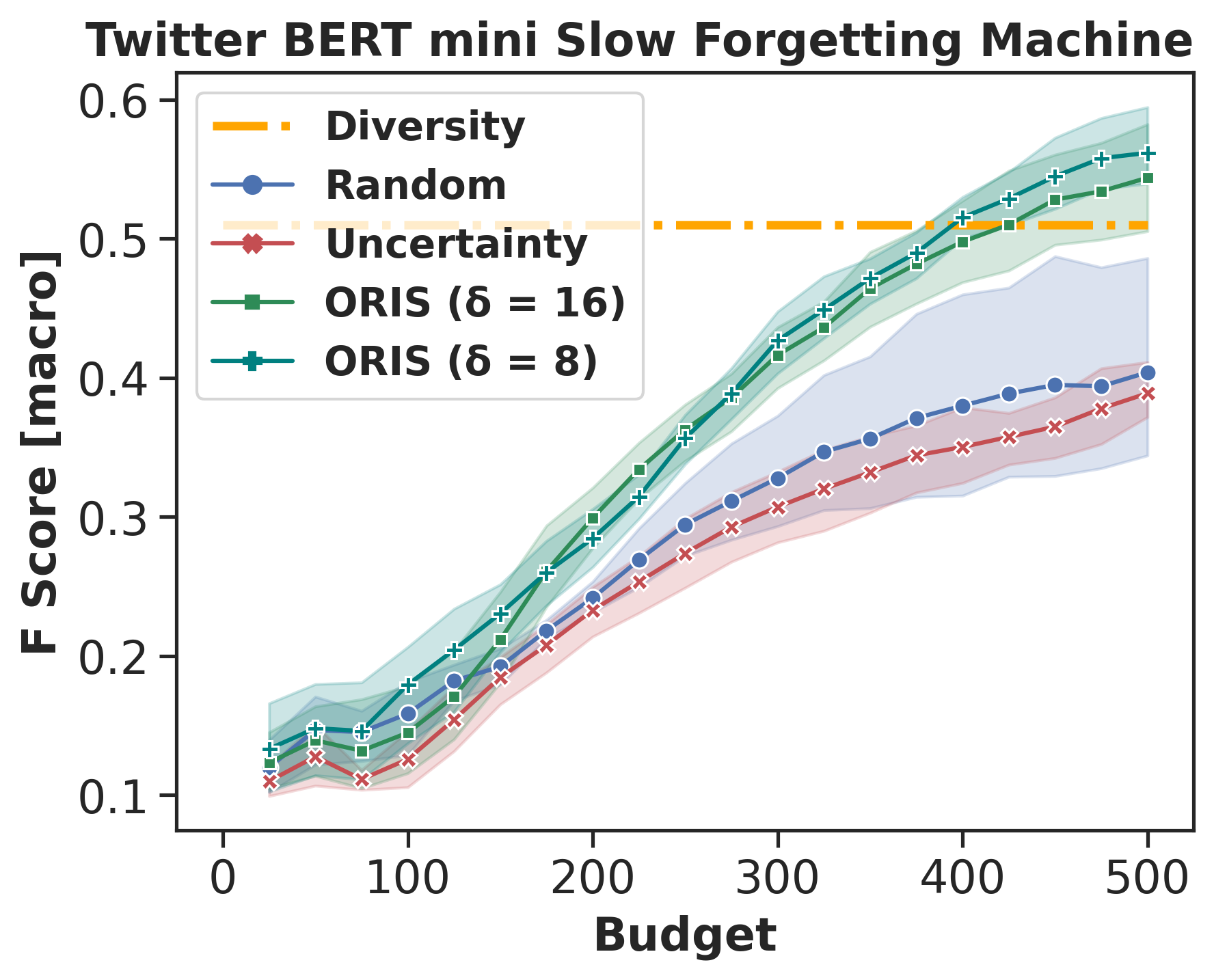}}
\subfigure{\includegraphics[width=0.23\textwidth]{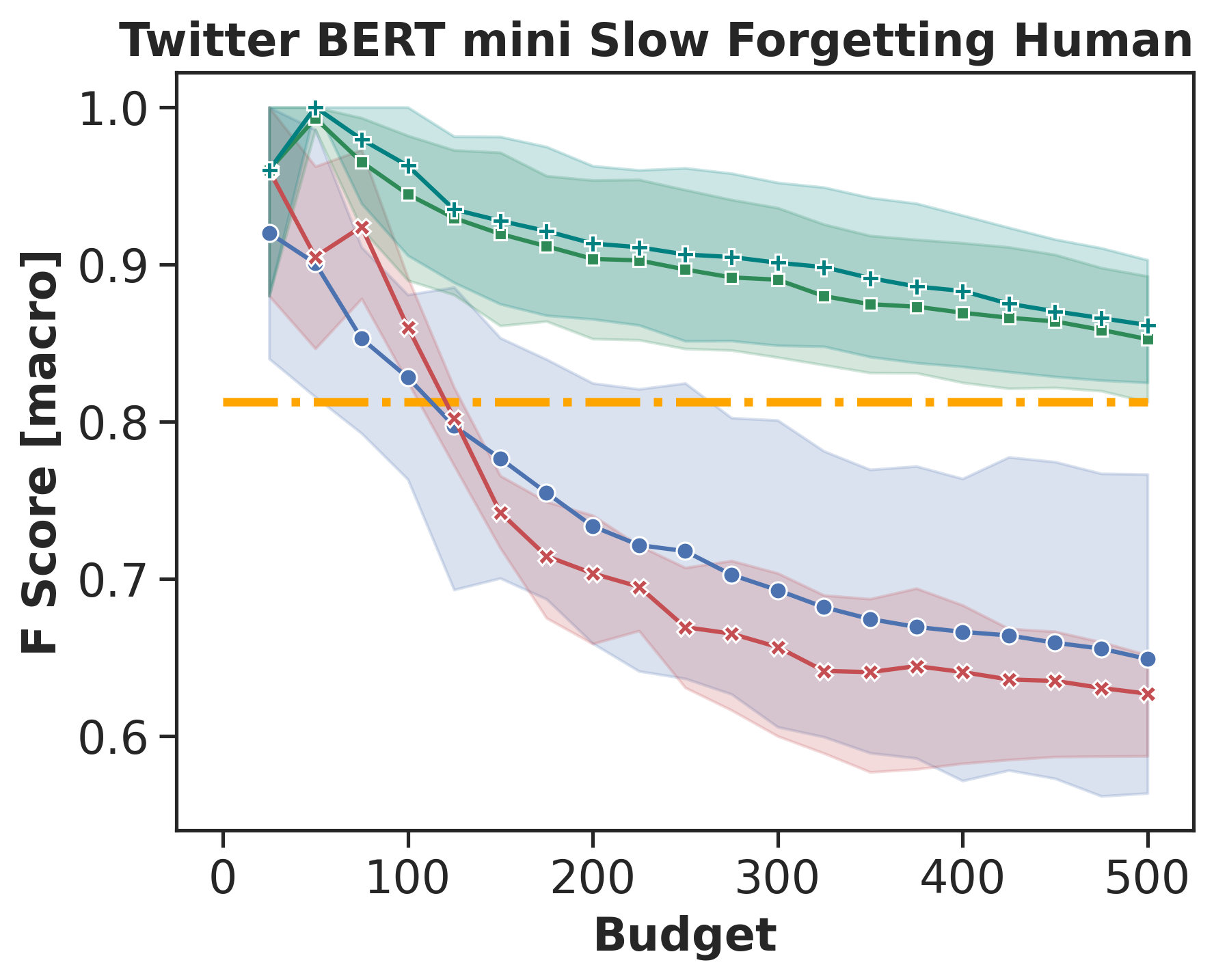}}
\subfigure{\includegraphics[width=0.23\textwidth]{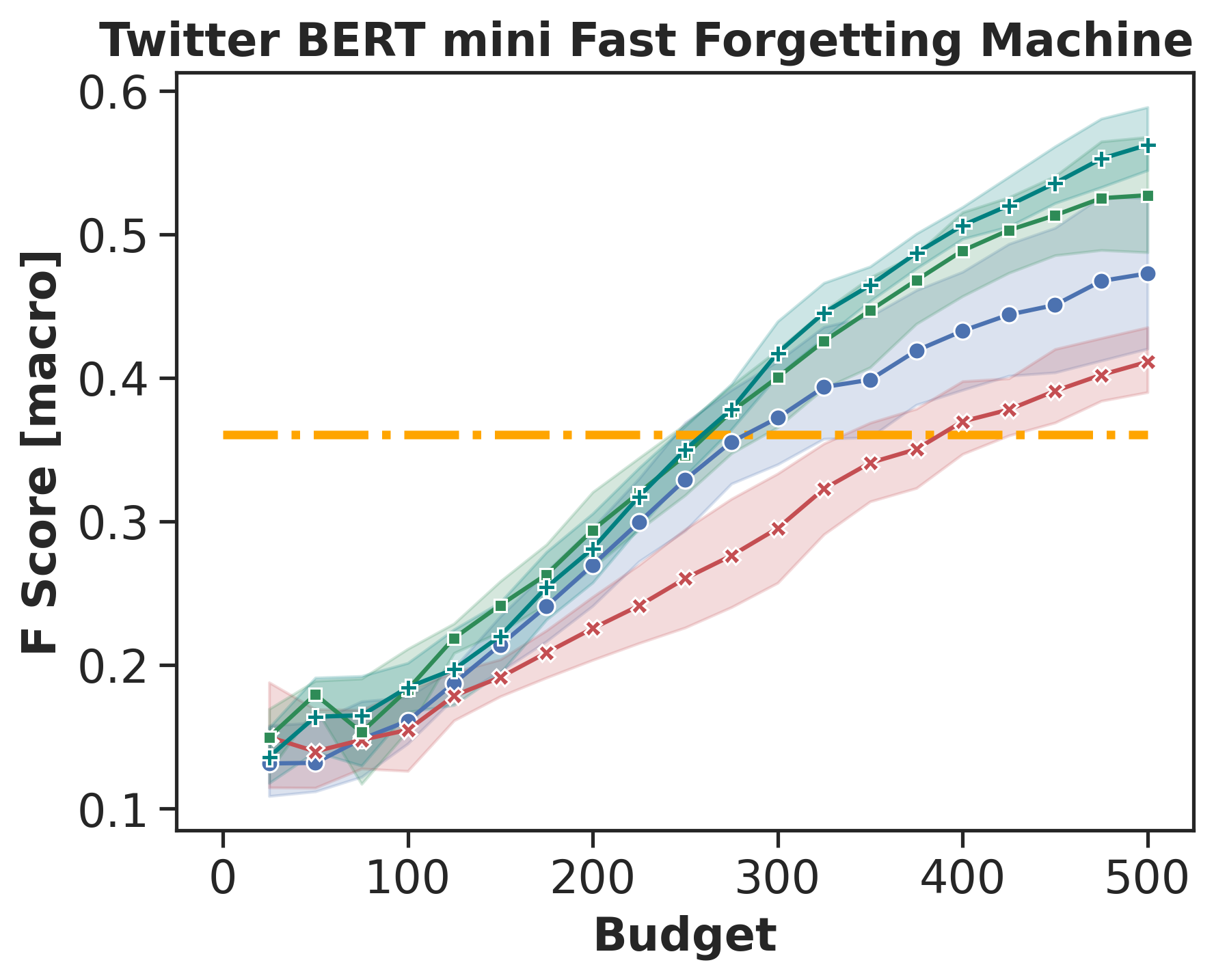}}
\subfigure{\includegraphics[width=0.23\textwidth]{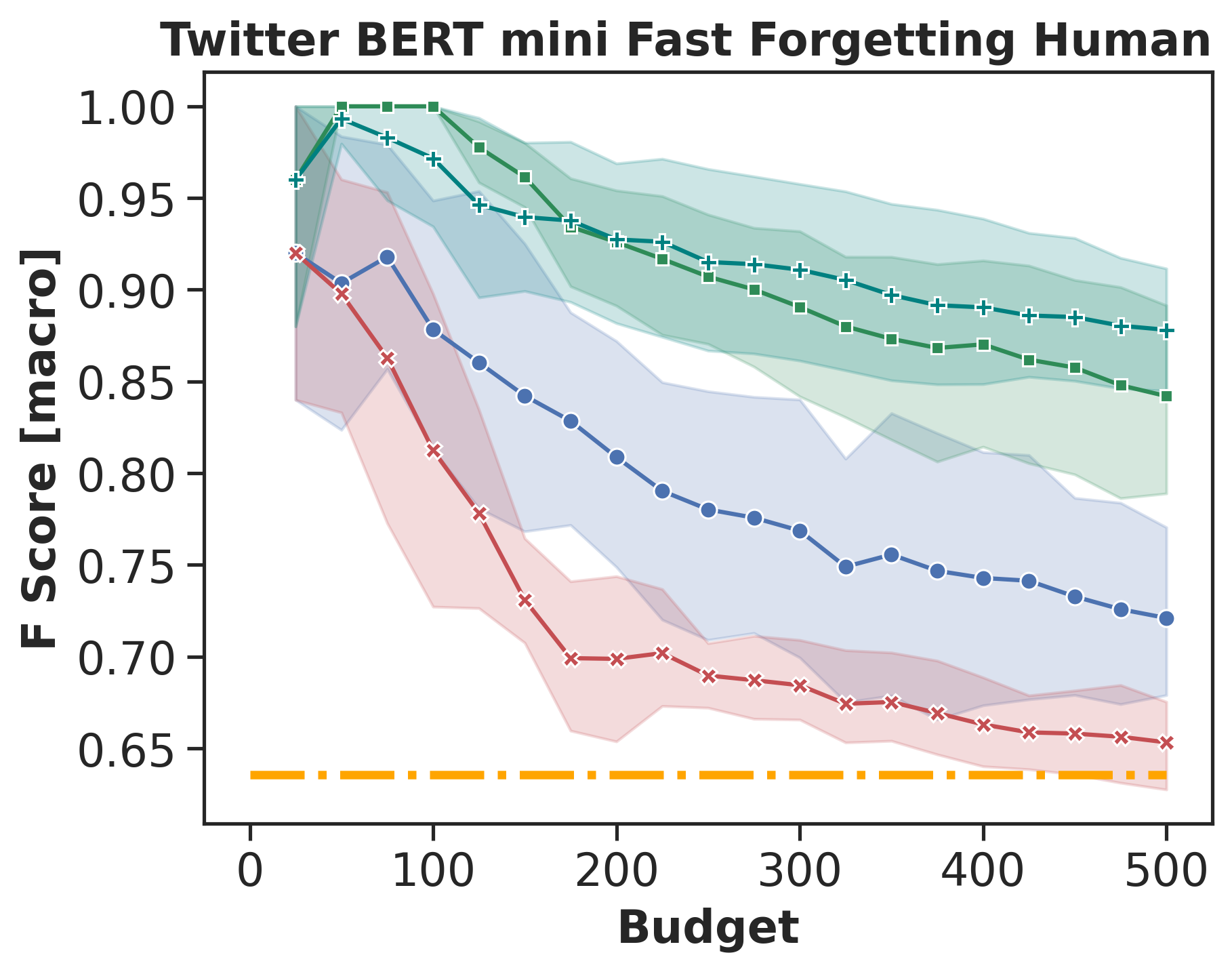}}
\subfigure{\includegraphics[width=0.23\textwidth]{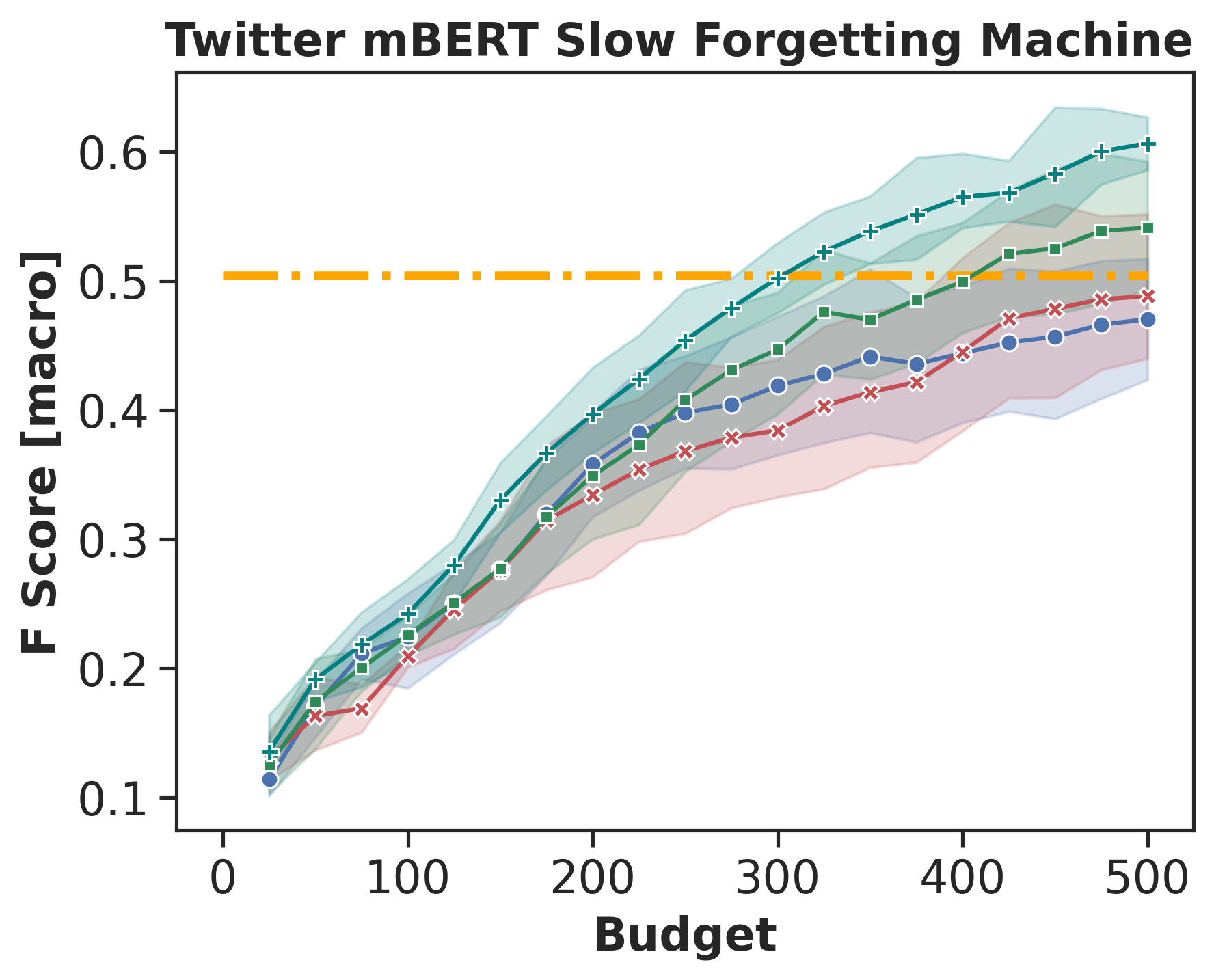}}
\subfigure{\includegraphics[width=0.23\textwidth]{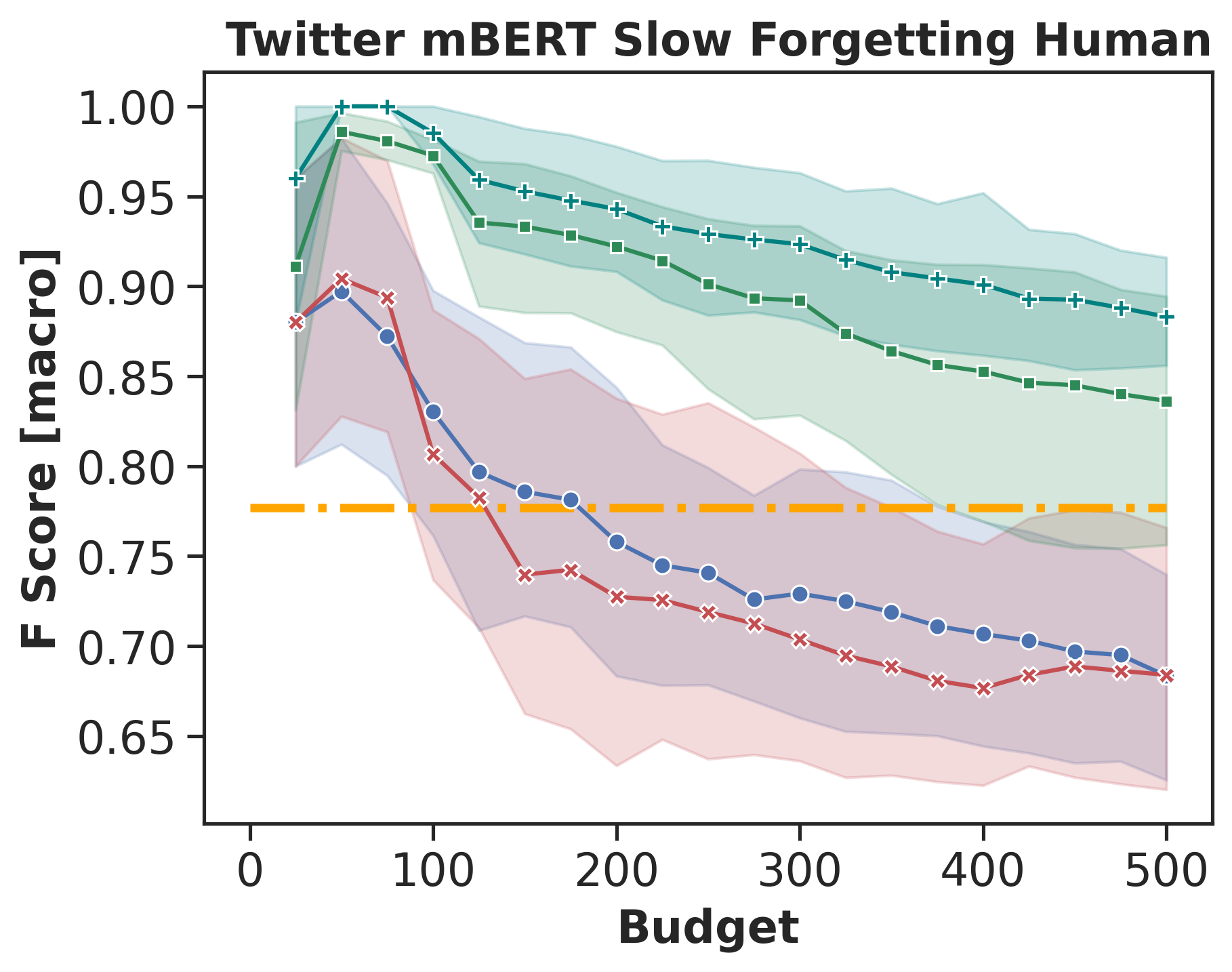}}
\subfigure{\includegraphics[width=0.23\textwidth]{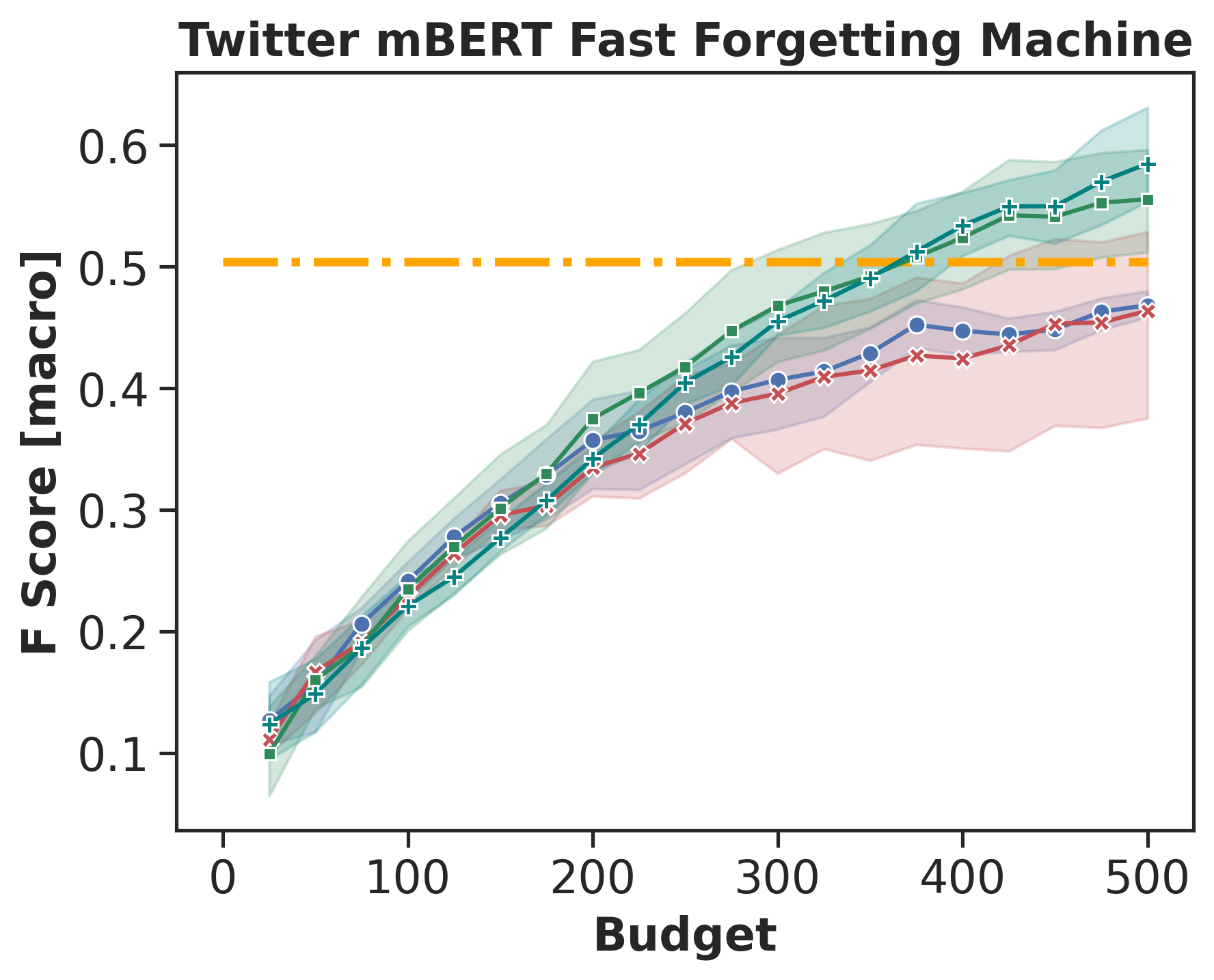}}
\subfigure{\includegraphics[width=0.23\textwidth]{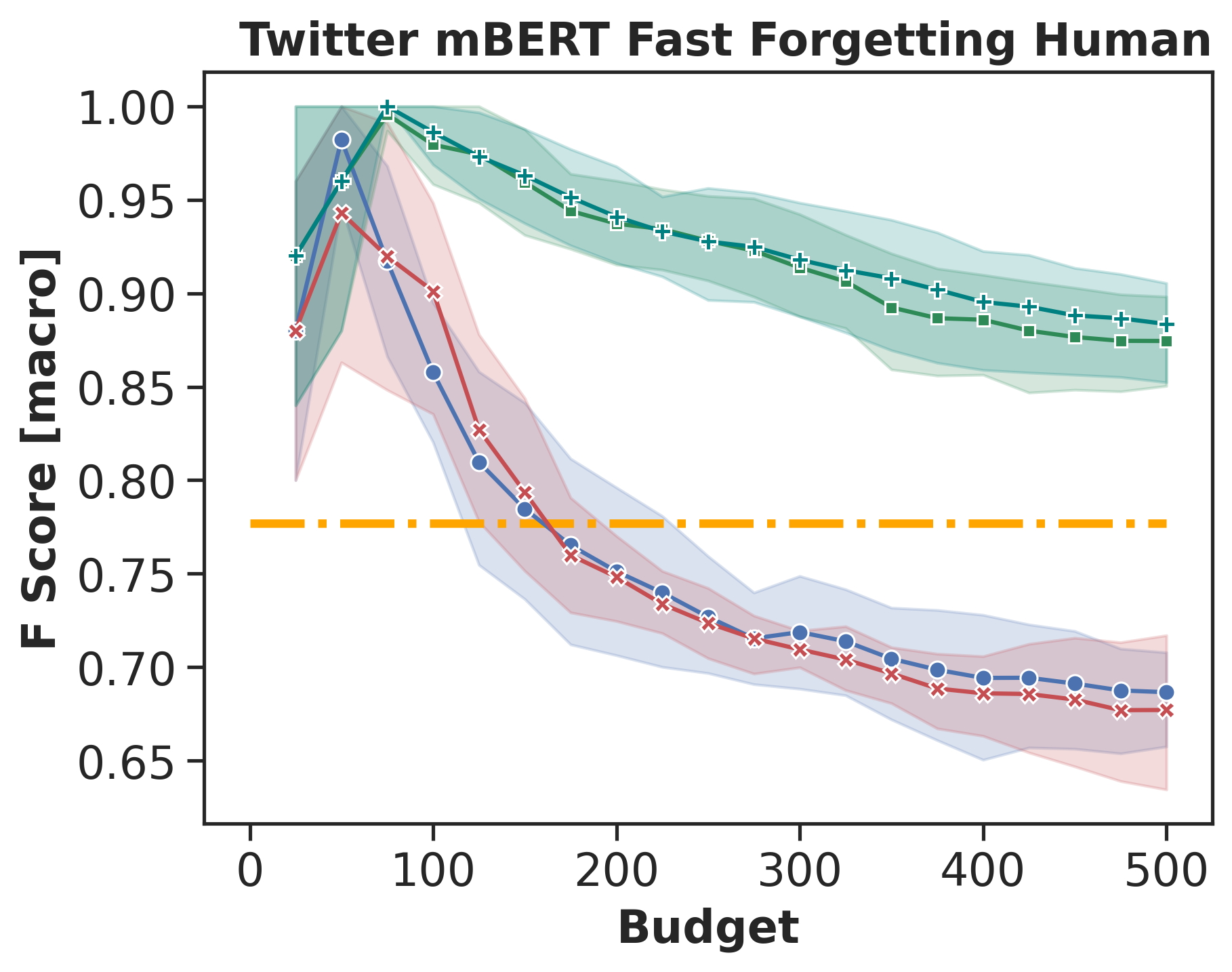}}
\subfigure{\includegraphics[width=0.23\textwidth]{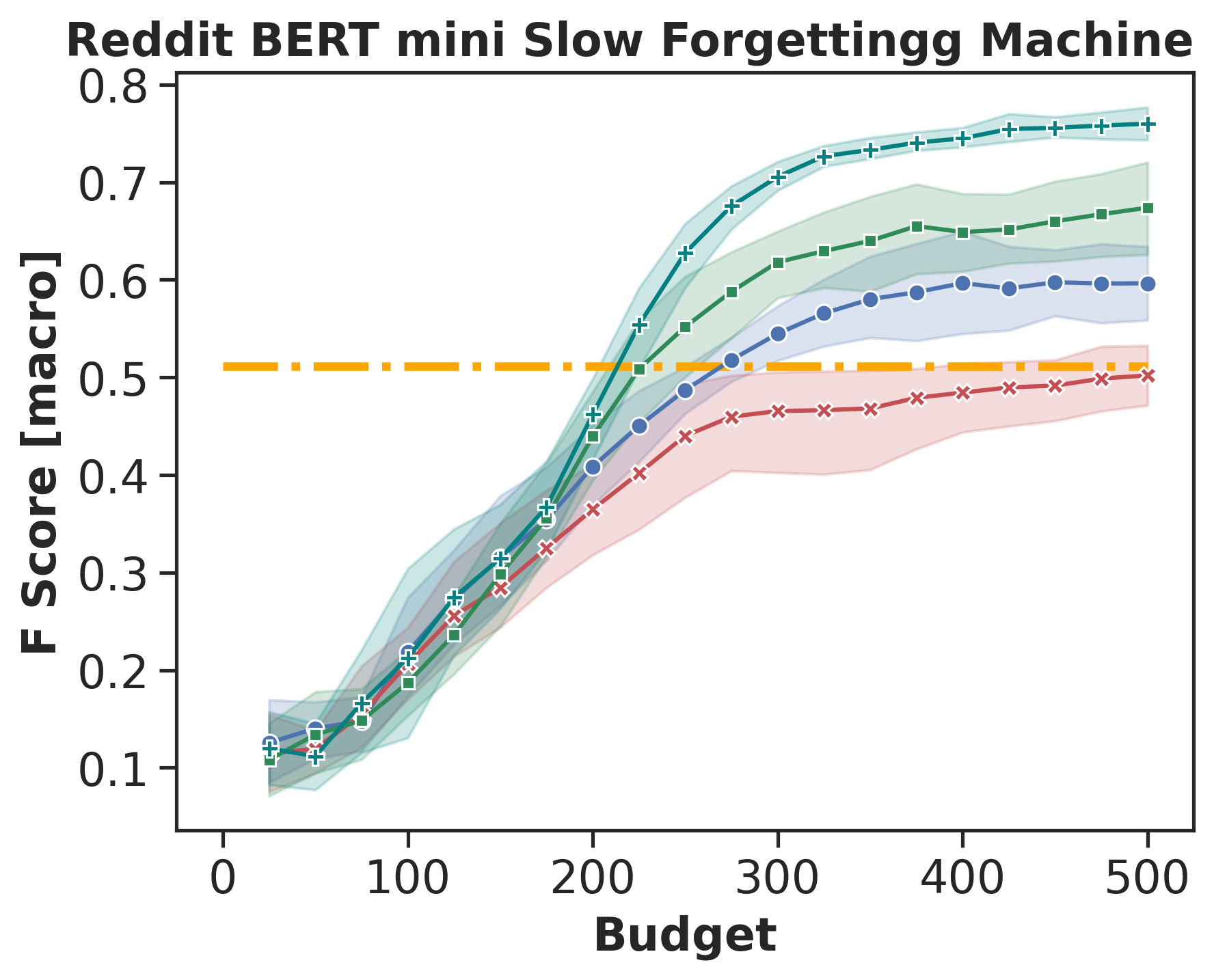}}
\subfigure{\includegraphics[width=0.23\textwidth]{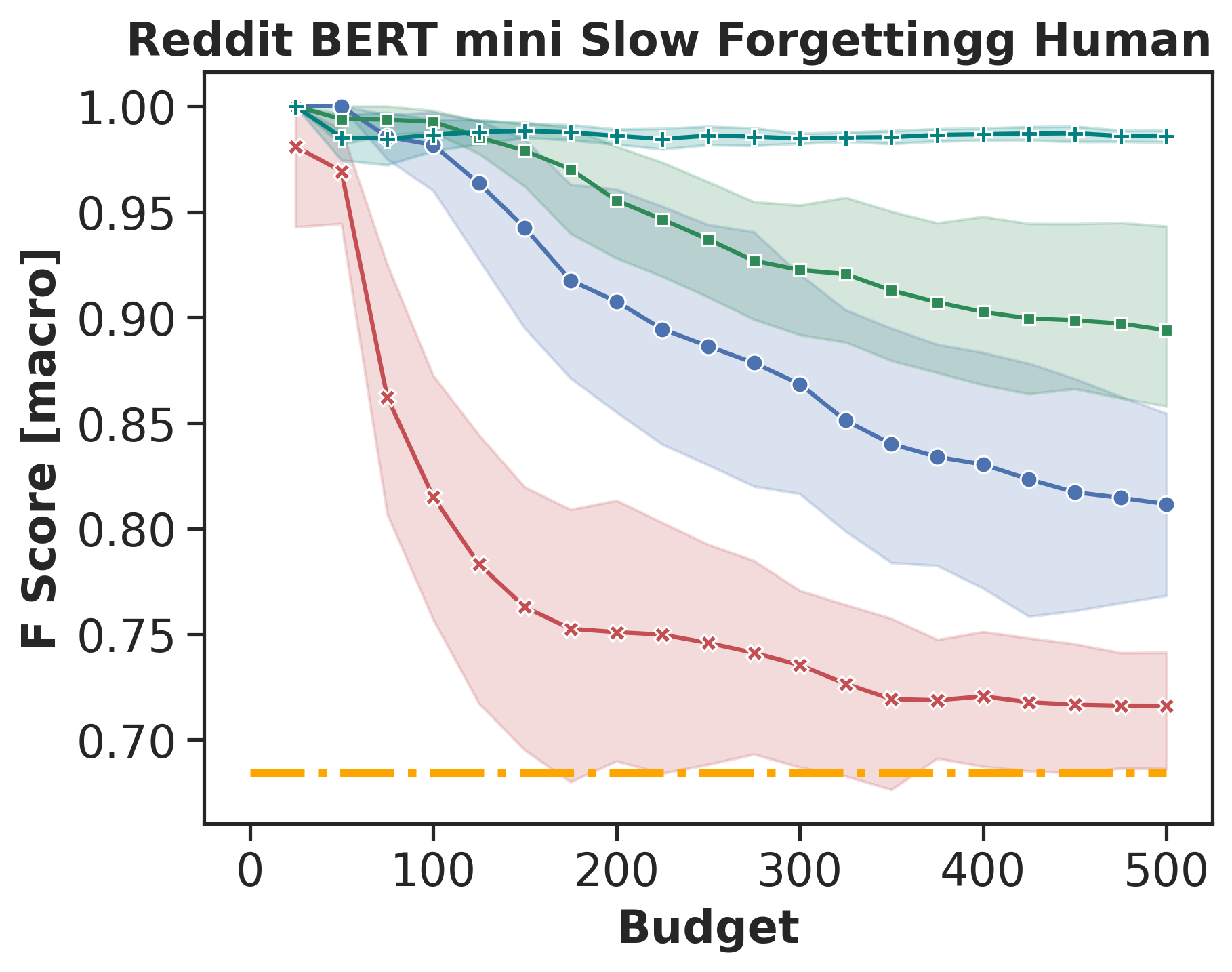}}
\subfigure{\includegraphics[width=0.23\textwidth]{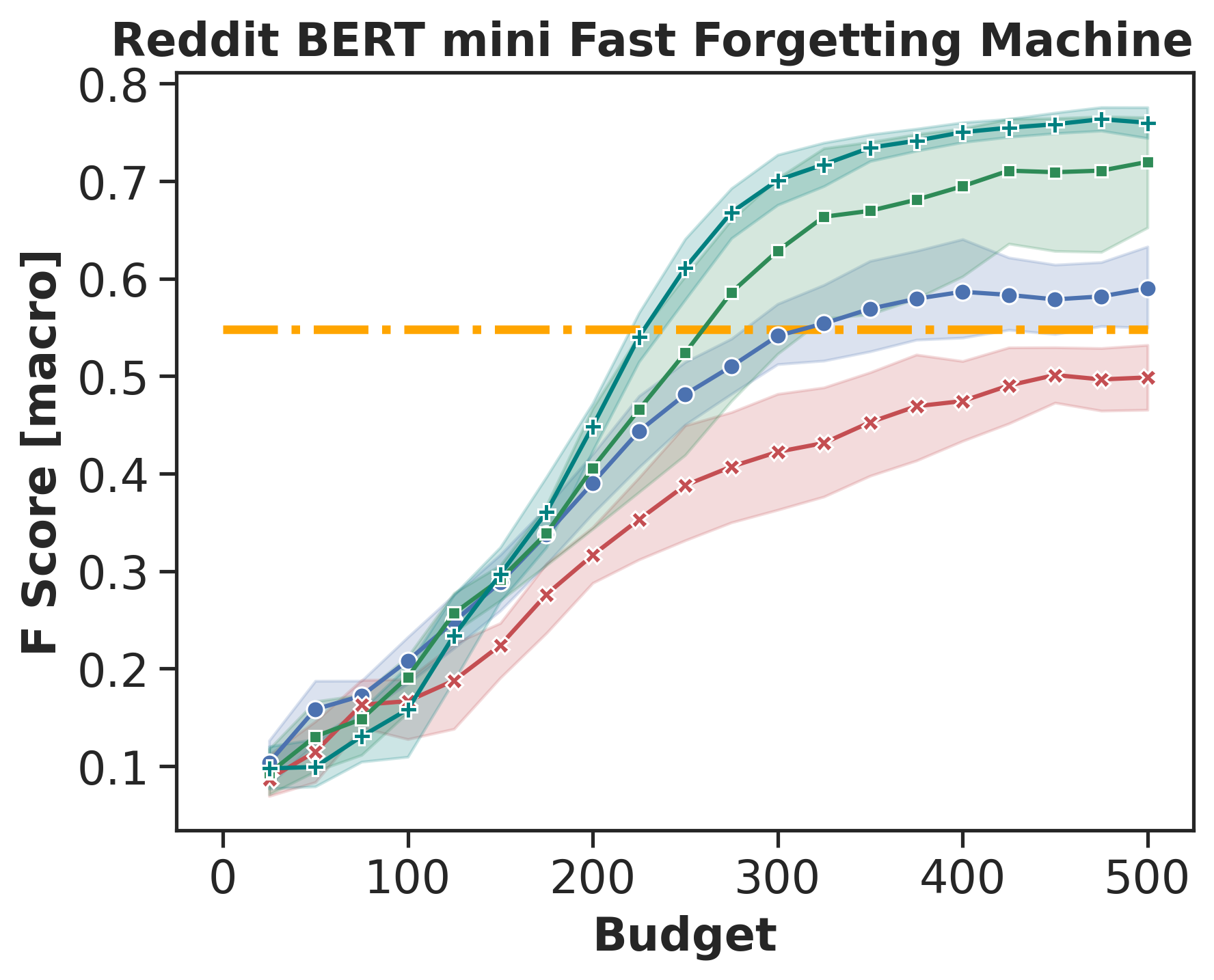}}
\subfigure{\includegraphics[width=0.23\textwidth]{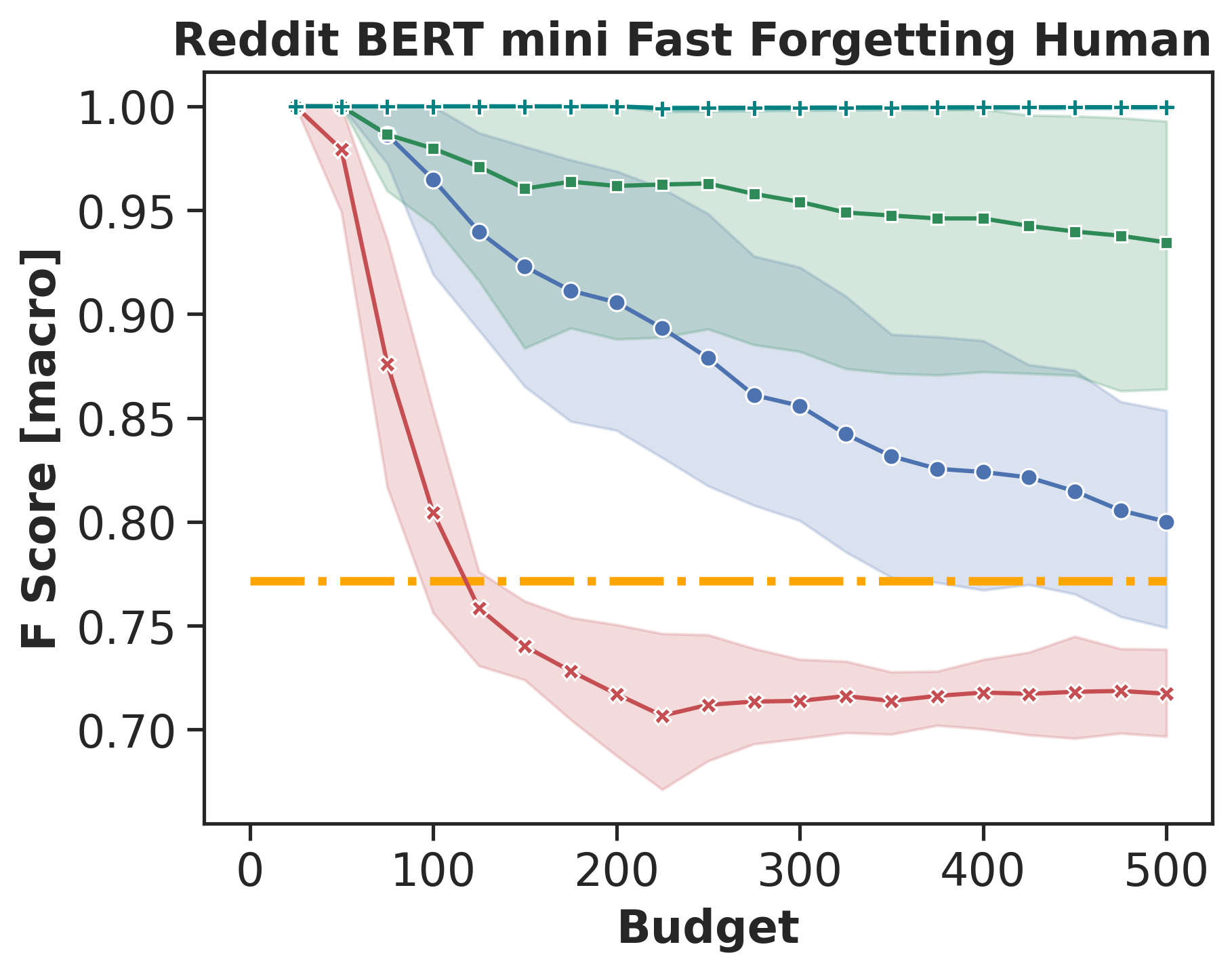}}
\subfigure{\includegraphics[width=0.23\textwidth]{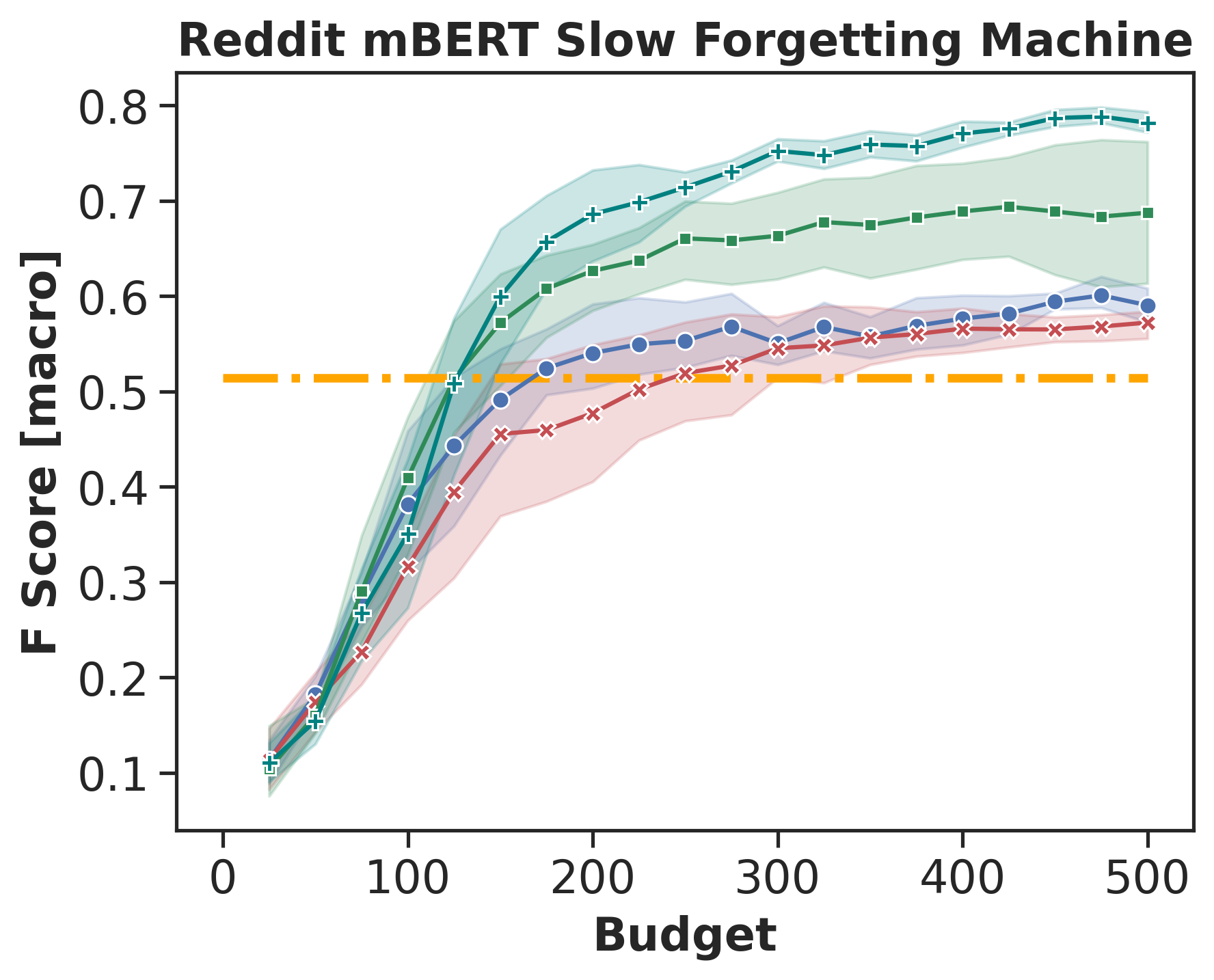}}
\subfigure{\includegraphics[width=0.23\textwidth]{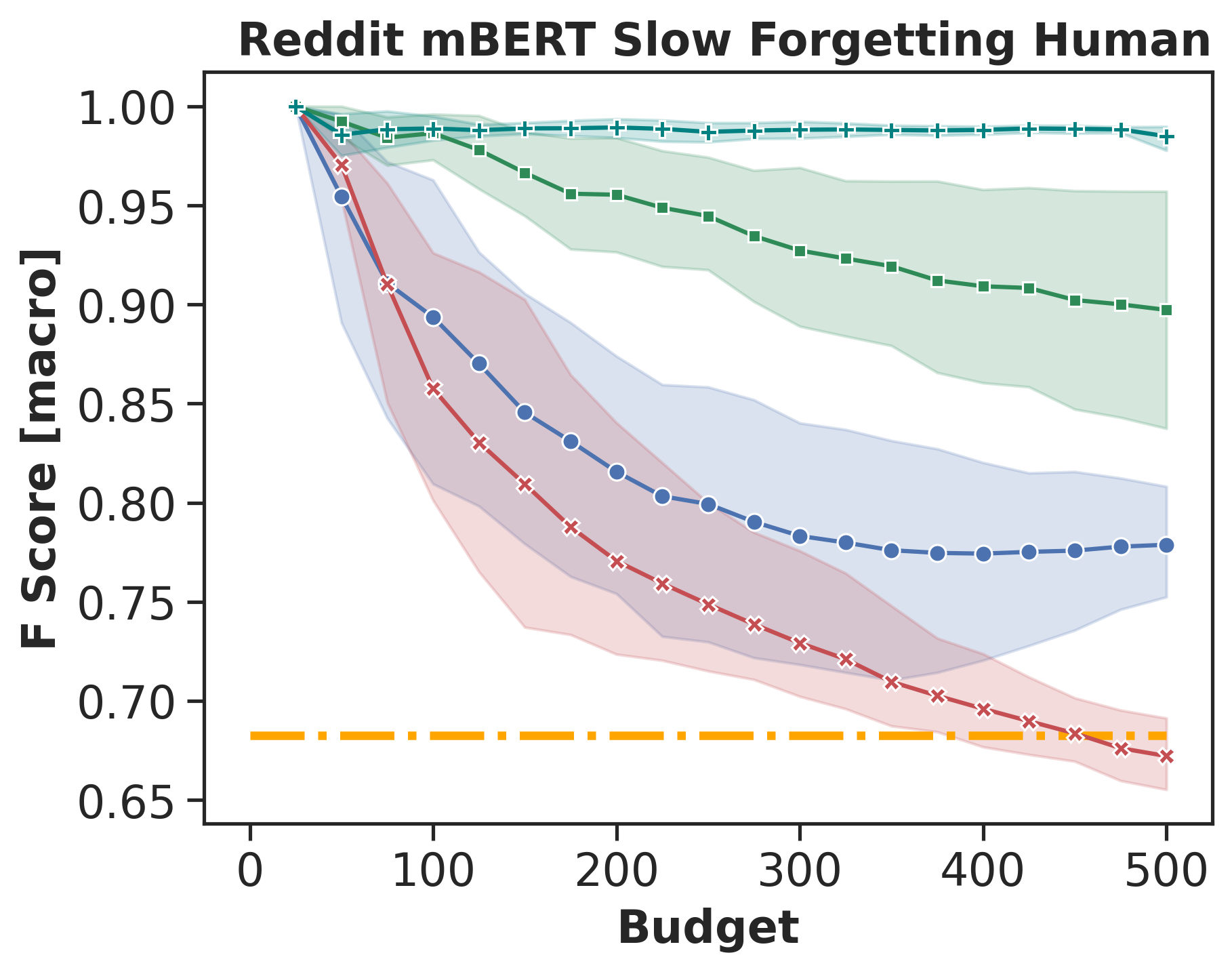}}
\subfigure{\includegraphics[width=0.23\textwidth]{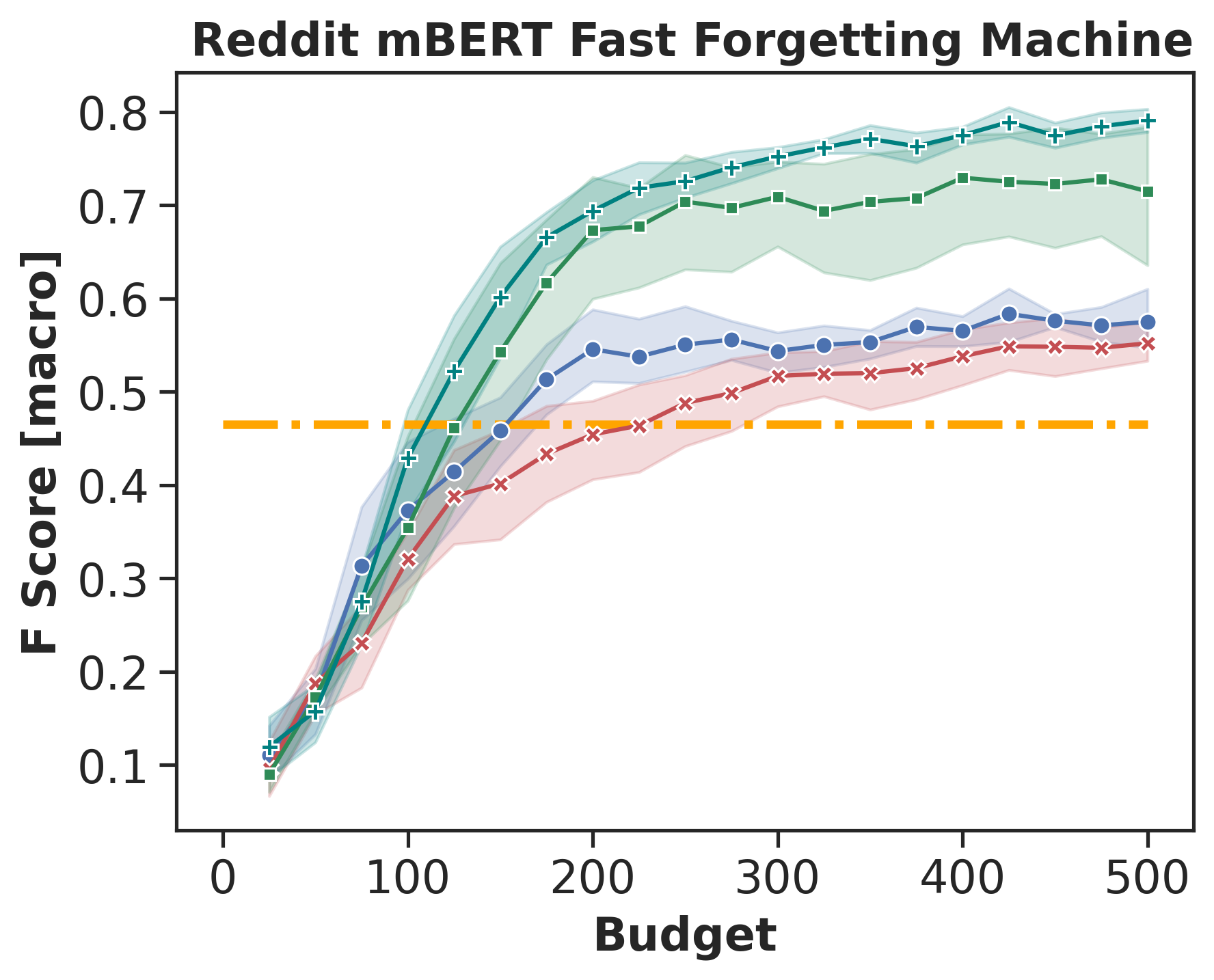}}
\subfigure{\includegraphics[width=0.23\textwidth]{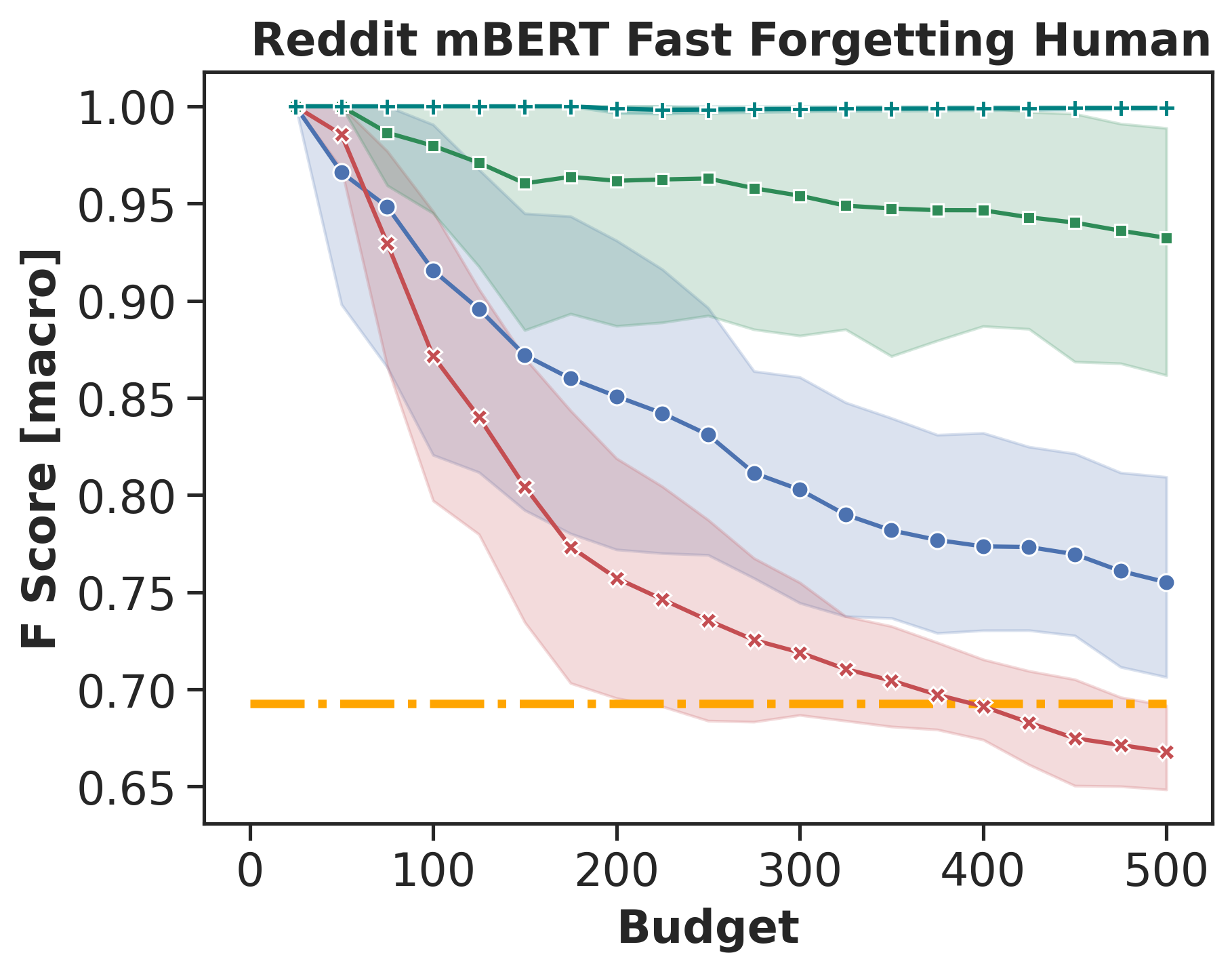}}
\caption{
Machine and Human performance comparison over budget exhausted for both Twitter and Reddit Dataset when using the BERT mini and mBERT model for fine-tuning.  
The shaded color area represents the 95\% confidence interval for the five runs. The blue and red shades represent random and uncertainty sampling, respectively. The offline diversity sampling is represented as a dashed yellow line along the x-axis. Both the green color shades represent the proposed ORIS experiments. The \textbf{$+$} mark represents the ORIS with $\delta=8$, and the {\tiny{$\blacksquare$}} mark represents the ORIS with $\delta=16$, respectively.
}
\label{fig:all_res}
\vspace{-4mm}
\end{figure*}
In the experiment, we answer four research questions. 
\begin{itemize}
    \item \textit{\textbf{RQ1:} How does the ORIS method reduce the impact of error-prone oracle and improve both ML model and human performance?} 
    \item \textit{\textbf{RQ2:} How effective is the ORIS method when transferred 
    across domains from Twitter to Reddit?}
   \item \textit{\textbf{RQ3:} What is the long-term effect on the performance of the ORIS method as the budget gets exhausted?}
    \item \textit{\textbf{RQ4:} How quickly and efficiently does the ORIS method perform?}
\end{itemize}

\subsection{Dataset}
\vspace{-2mm}
We implement the ORIS method for online active learning in emotion recognition, 
a well-studied complex natural language understanding task.
We use two famous social media datasets. The first dataset comprises of labeled Twitter posts with six emotions \cite{saravia-etal-2018-carer}. 
The second dataset comprises of labeled Reddit posts with manually annotated 27 emotions for (\textit{\textbf{RQ2}}) \cite{demszky-etal-2020-goemotions}. 
For our experiments, we filtered out the documents corresponding to only five common emotions classes \textit{(sadness, joy, surprise, anger, and fear)}. The Twitter dataset 
from the Huggingface Datasets~\footnote{\url{https://huggingface.co/datasets/dair-ai/emotion}} had two groups: split with train/validation/test distinction, and unsplit. We use the unsplit group minus the split data for DQN training. For both Twitter and Reddit data, we use train \& test splits for active learning demonstration and machine performance evaluation, respectively.  
Table~\ref{tab:dataset} shows the dataset distribution. 
Note that the datasets are highly imbalanced and are prone to memory decay during the labeling process. 
\subsection{Configuration}
\vspace{-2mm}
\begin{table*}[h!]
\caption{
Performance comparison for Twitter dataset 
in an active learning setting with budget $B=500$. 
Machine and human performance represent the mean and standard deviation of \textit{f1-macro} score of five random runs.
}

\label{table:res-tw}
\centering
\resizebox{\textwidth}{!}{\begin{tabular}{|l|c|c|c|c|c|c|c|c|}
\hline
\textbf{Experiment} & \multicolumn{4}{c|}{\textbf{BERT Mini}} & \multicolumn{4}{c|}{\textbf{mBERT}} \\
\cline{2-9}
& \multicolumn{2}{c|}{\textbf{Sigmoid}} & \multicolumn{2}{c|}{\textbf{Exponential}} & \multicolumn{2}{c|}{\textbf{Sigmoid}} & \multicolumn{2}{c|}{\textbf{Exponential}} \\
\cline{2-9}
& \textbf{Machine} & \textbf{Human} & \textbf{Machine} & \textbf{Human} & \textbf{Machine} & \textbf{Human} & \textbf{Machine} & \textbf{Human} \\
\hline
\textbf{Random} & 40.4 $\pm$ 8.6 & 64.9 $\pm$ 11.2 & 47.3 $\pm$ 6.3 & 72.1 $\pm$ 5.4 & 47.0 $\pm$ 5.8 & 68.3 $\pm$ 6.5 & 46.8 $\pm$ 1.2 & 68.7 $\pm$ 2.9 \\
\hline
\textbf{Uncertainty} & 38.9 $\pm$ 2.4 & 62.7 $\pm$ 3.9 & 41.1 $\pm$ 2.7 & 65.3 $\pm$ 2.7 & 48.8 $\pm$ 6.4 & 68.4 $\pm$ 8.0 & 46.4 $\pm$ 9.2 & 67.7 $\pm$ 4.6 \\
\hline
\textbf{Diversity} & 51.0 $\pm$ 0 & 81.2 $\pm$ 0 & 36.1 $\pm$ 0 & 63.5 $\pm$ 0 & 50.4 $\pm$ 0 & 77.7 $\pm$ 0 & 50.4 $\pm$ 0 & 77.7 $\pm$ 0 \\
\hline
\hline
\textbf{ORIS ($\delta = 16$)} & 54.4 $\pm$ 4.4 & 85.2 $\pm$ 4.6 & 52.7 $\pm$ 4.3 & 84.2 $\pm$ 6.1 & 54.1 $\pm$ 6.0 & 83.6 $\pm$ 8.2 & 55.6 $\pm$ 4.8 & 87.4 $\pm$ 2.9 \\
\hline
\textbf{ORIS ($\delta = 8$)} & \textbf{56.2 $\pm$ 3.2} & \textbf{86.1 $\pm$ 4.6} & \textbf{56.2 $\pm$ 2.7} & \textbf{87.8 $\pm$ 4.0} & \textbf{60.6 $\pm$ 2.4} & \textbf{88.3 $\pm$ 3.6} & \textbf{58.5 $\pm$ 4.5} & \textbf{88.4 $\pm$ 3.1} \\
\hline
\end{tabular}
}
\vspace{-2mm}
\end{table*}
\subsubsection{Hyperparameter tuning} 
The ORIS method has several parameters, outlined here. To train the DQN agent, we run for $E_{max} = 10,000$ episodes with budget $B = 500$. We kept the replay buffer $\mathcal{R}$ size to $50,000$. The DQN architecture shown in Eq~\ref{eqn:dqn-network} has dense layer sizes of 256, 256, and 2, respectively. The last layer represents the policy for output actions: pick or discard. It takes input from the state defined in Eq~\ref{eqn:state}. We use FastText\footnote{\url{https://fasttext.cc/docs/en/aligned-vectors.html}} word embeddings for document representation~\cite{bojanowski2017enriching,joulin2018loss}. 
To compute the embeddings, consider the current streaming document $d_t$ at time $t$ as set of words $[w_1, w_2, ..., w_n]$ of length $n$, we compute embeddings as $emb_t = \frac{\sum_{w \, \epsilon \, d_t} emb^{w}}{n}$. 
The exploration rate $\epsilon$ is initialized with $0.9$ and exponentially decays to $0.05$ with the decay rate of $0.0005$. 
The discount factor $\gamma$ is set to $0.99$. The mini-batch transition size to sample from 
the replay buffer $\mathcal{R}$ is set to $512$. We update the target network with factor $\tau = 0.005$. The DQN minimizes the smooth L1 loss with a learning rate of $1e-4$. To compute the reward in Eq~\ref{eqn:reward}, we use the past $m = 10$ memory to get the Inclusivity score. The small positive reward for exploration $\lambda$ is set to $0.01$. The parameter $\rho$ in Eq~\ref{eqn:reward} is set to $5$. During inference, we use two large language models as ML models to fine-tune in an online active learning setting: \textit{BERT-Mini}~\cite{bhargava2021generalization, turc2020wellread} and \textit{mBERT Base}~\cite{devlin_bert_2019}. For both pretrained models, the embedding layer is frozen during fine-tuning. The model updates with update frequency $f = 25$ till the budget $B = 500$ is exhausted. For each fine-tuning 
, we keep the batch size to 8, the learning rate to $2e-5$, the number of epochs for training to $5$, and weight decay to $0.01$. 
\subsubsection{Experimental Setting} 
We propose two variations of ORIS with the reward activation 
$\delta$. 
\textit{ORIS ($\delta = 16$):} In this experiment, the reward is heavily deactivated for low scores of inclusivity, e.g., 
the inclusivity score of $< 0.8$ has a negligible reward.  
\textit{ORIS ($\delta = 8$):} In this,  
the reward quickly activates compared to the previous experiment. However, the lower score of inclusivity has a negligible reward. 
Furthermore, 
we mimic two oracles with different parameterized memory decay explained in the section~\ref{sec:pre_mem}: 

\noindent \textit{- Sigmoid [Slow] Forgetting:} We use the sigmoid-based decay function (\textit{c.f.} Eq~\ref{eqn:err_sig}) with $\alpha=0.3$ and $\beta = 9$. 

\noindent \textit{- Exponential [Fast] Forgetting:} We use the exponential-based decay function (\textit{c.f.} Eq~\ref{eqn:err_exp}) with $\alpha=0.6$ and $\beta=-19$. 
We report the results of five runs with different random ordering to simulate 
streaming 
settings. 

\begin{table*}[h] 
\caption{Performance comparison for Reddit dataset 
in an active learning setting with budget $B=500$. 
Machine and human performances  represent the mean and standard deviation of \textit{f1-macro} score of five random runs. 
}
\label{tab:res-rd}
\centering
\resizebox{\textwidth}{!}{\begin{tabular}{|l|c|c|c|c|c|c|c|c|}
\hline
\textbf{Experiment} & \multicolumn{4}{c|}{\textbf{BERT Mini}} & \multicolumn{4}{c|}{\textbf{mBERT}} \\
\cline{2-9}
& \multicolumn{2}{c|}{\textbf{Sigmoid}} & \multicolumn{2}{c|}{\textbf{Exponential}} & \multicolumn{2}{c|}{\textbf{Sigmoid}} & \multicolumn{2}{c|}{\textbf{Exponential}} \\
\cline{2-9}
& \textbf{Machine} & \textbf{Human} & \textbf{Machine} & \textbf{Human} & \textbf{Machine} & \textbf{Human} & \textbf{Machine} & \textbf{Human} \\
\hline
\textbf{Random} & 59.6 $\pm$ 4.5 & 81.2 $\pm$ 5.3 & 59.0 $\pm$ 4.9 & 80.0 $\pm$ 6.1 & 59.0 $\pm$ 2.1 & 77.9 $\pm$ 3.1 & 57.5 $\pm$ 3.8 & 75.5 $\pm$ 5.9\\
\hline
\textbf{Uncertainty} & 50.2 $\pm$ 3.5 & 71.6 $\pm$ 3.1 & 49.9 $\pm$ 3.9 & 71.7 $\pm$ 2.4 & 57.2 $\pm$ 1.6 & 67.2 $\pm$ 2.1 & 55.2 $\pm$ 2.5 & 66.8 $\pm$ 2.4\\
\hline
\textbf{Diversity} & 51.1 $\pm$ 0 & 68.5 $\pm$ 0 & 54.8 $\pm$ 0 & 77.2 $\pm$ 0 & 51.4 $\pm$ 0 & 68.3 $\pm$ 0 & 46.5 $\pm$ 0 & 69.3 $\pm$ 0\\
\hline
\hline
\textbf{ORIS ($\delta = 16$)} & 67.4 $\pm$ 5.6 & 89.4 $\pm$ 4.9 & 72.0 $\pm$ 6.8 & 93.5 $\pm$ 7.3 & 68.7 $\pm$ 8.4 & 89.7 $\pm$ 7.0 & 71.5 $\pm$ 8.4 & 93.2 $\pm$ 7.2\\
\hline
\textbf{ORIS ($\delta = 8$)} & \textbf{76.0 $\pm$ 1.9} & \textbf{98.6 $\pm$ 0.3} & \textbf{76.0 $\pm$ 1.8} & \textbf{100.0 $\pm$ 0.1} & \textbf{78.2 $\pm$ 1.3} & \textbf{98.5 $\pm$ 0.7} & \textbf{79.1 $\pm$ 1.5} & \textbf{99.9 $\pm$ 0.1}\\
\hline
\end{tabular}
}
\vspace{-2mm}
\end{table*}

\subsection{Baseline Techniques} 
\vspace{-2mm}
We compared the proposed ORIS method to several traditional online active learning. Since our approach to computing inclusivity is loosely based on diversity sampling idea, we also include additional experiments of offline diversity-sampling-based active learning~\cite{bodo_active_2011, hu_off_2010}. To our knowledge, we could not find any online diversity sampling technique. Here is the list of different baseline methods:

\noindent \textit{1. [Online] \textbf{Random} Sampling: 
} The agent 
    randomly decides to pick or discard the current instance. It is highly dependent on data distribution.
    
\noindent    \textit{2. [Online] \textbf{Uncertainty} Sampling: 
} The agent 
    utilizes the ML model's confidence to compute entropy and use a threshold for decision, which we dynamically reduce 
    with the budget utilization~\cite{wang2016cost}.
    
\noindent    \textit{3. [Offline] \textbf{Diversity} Sampling:  
} The agent uses offline agglomerative clustering~\cite{bodo_active_2011} 
to create $B$ clusters and sample $B$ documents closest to the center.   

\subsection{Evaluation Metrics} 
\vspace{-2mm}
We report \textit{f1-macro} scores for both 
machine and human performance. 
For machine performance, we rely on the independent test data. At every update frequency $f$, when the ML model is trained with labels provided by the oracle, we compute the \textit{f1} scores for each class $c$ as 
$f1_c$ and then, \textit{f1-macro} score. 
%
For human performance, we rely on the labels provided by the oracle for the selected documents. 
To compute \textit{f1} scores for each class $c$ as 
$f1_c$,  
true positive 
means    
when the oracle labeled the document with class $c$ correctly. Similarly, 
false positive 
means   
when the oracle erroneously labeled the document as class $c$ but originally belonged to another class, 
and false negative  
means   
when the oracle erroneously labeled the document that originally belonged to class $c$ to another class. 
Finally, we compute the \textit{f1-macro} score. 
\section{Results \& Discussion}
To answer the \textbf{\textit{RQs}} stated in Section~\ref{sec:exp}, we report the results of both the evaluation metrics for the online active learning implementation for two datasets. 
Table~\ref{table:res-tw} and~\ref{tab:res-rd} show the final performance when the budget $B=500$ is exhausted for each 
experiment. 
We divide the results into four parts. In the first part, we explore the final performance of active learning in the error-prone environment for Twitter data, and compare the ORIS method with baselines. In the second part, we focus on the effect of ORIS method when used for active learning on cross-domain Reddit data.  
In the third part, we focus on long-term effect of the proposed ORIS method. Finally, in the fourth part, we compare the inference speed of the different active learning demonstrations.  

\subsection{Performance comparison with ORIS (\textit{\textbf{RQ1}})}
\vspace{-2mm}
Table~\ref{table:res-tw} shows the results of experiments conducted with the Twitter dataset. We observe that the final human performance is significantly lower in the baselines compared to the proposed ORIS method. We also observe that there is a good machine performance degradation even in the robust large language model like BERT Mini and mBERT if there is a significant amount of labeling error present. We observe the degradation in \textit{f1-macro} score as low as $40.4\%$ for random sampling compared to the maximum of $56.2\%$ for the proposed ORIS method with $\delta=8$ when trained with an equal number of documents ($B=500$). Similarly, the difference can reach from $46.8\%$ for random sampling compared to the maximum of $60.6\%$ for the proposed ORIS sampling with $\delta=8$ when fine-tuning an mBERT model. Moreover, within the baseline strategies, diversity sampling has better human performance than random or uncertainty-based sampling showing that diversity sampling can explore and extract inclusive samples to reduce human errors of the oracle, further improving the machine performance. We observe that random sampling has high variance since it is based on the distribution of the incoming document stream. 

Furthermore, both the proposed ORIS methods performed significantly better than the baselines 
in terms of both the human performance of error-prone oracle and the machine performance of the two BERT models. However, 
the experiments with $\delta=8$ have higher and more stable performance than $\delta=16$. We believe that with $\delta=8$, the reward values are more informative for robust and efficient DQN training.

\subsection{Domain transfer effect comparison (\textit{\textbf{RQ2}})}
\vspace{-2mm}
Table~\ref{tab:res-rd} shows the results of experiments conducted on the Reddit emotion dataset. 
Since the Reddit dataset is cross-domain for the proposed ORIS method and has reliable manual labels, the results demonstrate the significance of our approach for a more general setting. Similar to Table~\ref{table:res-tw}, we observe that the proposed ORIS method improved both BERT Mini and mBERT models compared to all the baselines, including the offline diversity sampling. We observe that the offline diversity sampling had lower human performance compared to other baselines, contrary to what we observe in the Twitter dataset. We believe it could be due to the smaller sample size 
active learning sampling. Since the Reddit training data is $\approx 4$ times less than the Twitter data, the offline diverse sampling may represent diversity among the same class. Hence, the sampled documents were still prone to memory decay-based errors from the error-prone oracle. However, the 
ORIS method outperformed all baselines in this dataset. The ORIS method with $\delta=8$ achieved an \textit{f1-macro} score of $76.0\%$ on the BERT Mini model and $79.0\%$ on the mBERT model. We observe the effect of degradation in human performance in the baselines that do not take inclusivity into consideration. The human performance has reached as low as $66.8\%$ \textit{f1-macro} score in the baseline uncertainty sampling. While the 
ORIS method 
reached $99.9\%$ \textit{f1-macro} in human performance, which signifies that the oracle, despite being prone to memory decay, was able to make close to zero errors in labeling.  

\subsection{Long term effect comparison (\textit{\textbf{RQ3}})}
\vspace{-2mm}
In this section, we look into the performance progress in our experiments through Fig.~\ref{fig:all_res}. Collectively, we observe that the proposed ORIS method starts gaining significant performance improvement in machine performance with budget utilization of as low as 150 documents. However, the human performance gain is visible when the budget utilization is as low as 50 documents. It shows that even if the effect of labeling errors does not directly correlate with machine performance, it will eventually hamper the machine's performance in the long run. 
Based on the error-prone oracle design as defined in Section 3.2, the human performance is highest at the beginning as the classes are just learned. 
However, we observe that human performance degrades at a much higher acceleration for baselines as compared to the proposed ORIS method. We observe that uncertainty sampling is most affected by human performance compared to other baselines since it does not consider the distribution of the incoming data. Even though the uncertainty sampling chose informative samples to improve machine performance, due to the high erroneous labels provided by the oracle, it does not improve the machine performance compared to the other baselines. 
There are several other hyperparameters that can be tweaked for better performance, which we will study in the future. 

\subsection{Inference speed comparison (\textit{\textbf{RQ4}})}
\vspace{-2mm}
\begin{table}[ht]
  \centering
  \caption{Average inference duration of active learning experiments. Each value represents the average duration 
  in H:MM:SS. 
  }
  \label{tab:speed}
  \begin{tabular}{|c|c|c|c|c|}
    \hline
    \textbf{Experiment} & \multicolumn{2}{c|}{\textbf{Twitter}} & \multicolumn{2}{c|}{\textbf{Reddit}} \\ \hline
     & BERT Mini & mBERT & BERT Mini & mBERT \\
    \hline
    \textbf{Random} & 0:01:00 & 0:04:00 & 0:00:47 & 0:02:37 \\ \hline
    \textbf{Uncertainty} & 0:01:04 & 0:04:32 & 0:00:50 & 0:03:06 \\ \hline
    \textbf{Diversity} & 0:01:09 & 0:04:50 & 0:00:46 & 0:02:40 \\ \hline \hline
    \textbf{ORIS ($\delta = 16$)} & 0:00:55 & 0:04:03 & 0:00:45 & 0:02:43 \\ \hline
    \textbf{ORIS ($\delta = 8$)} & 0:01:00 & 0:04:01 & 0:00:47 & 0:02:37 \\ \hline
  \end{tabular}
  \vspace{-2mm}
\end{table}

Table~\ref{tab:speed} shows the average experiment duration when completing the sampling of $B$ documents and the performance evaluation at every update frequency $f$. For the diversity sampling, we do not average
as there was only one run per experiment. All experiments ran on Linux machine with four CPU cores, 32GB memory, and one Nvidia A100 GPU with 40GB memory for GPU computation (for BERT model fine-tuning and DQN inference).  We observe that the experiments with  mBERT fine-tuning took significantly more time than the BERT Mini fine-tuning. It is evident because the size of the mBERT model is larger than the BERT mini. Moreover, since Twitter data exploration was higher than Reddit data, it took a long time to finish the experiments. We observe that both of our proposed ORIS methods are faster than uncertainty or diversity sampling in all experimental settings. It could be because calculating the confidence of each documents in uncertainty sampling can be time-consuming. Moreover, diversity sampling requires the computation of the clustering algorithm for entire data beforehand, which is, again, time-consuming. And the proposed ORIS method performs similarly to random sampling, which makes the sampling decision in near real-time. The only time-consuming part of the experiments was fine-tuning of BERT model, which was common in all experiments. Overall, the results demonstrate that the proposed method, ORIS, is highly effective for online active learning in real-world error-prone settings.
\section{Conclusion \& Future Work}
\vspace{-1mm}
In this paper, we address the challenge of human memory decay that causes errors in data labeling process, by designing a novel and efficient online active learning-based streaming analytics system. 
We presented ORIS, a novel method that addresses both human and machine performance challenges  
through a reinforcement learning problem formulation. We introduced a novel Inclusivity factor in designing the reward of a Deep Q-Network. We evaluated the 
ORIS method on emotion recognition tasks with 
traditional active learning baselines and 
analyzed both human and machine performance. We observe that the traditional baselines are prone to human errors when we use the slip-based error-prone oracle. Whereas the 
ORIS method reduces human errors by 
inclusively sampling 
documents and thus, improving the ML 
model performance. We observe this performance improvement on two datasets: Twitter and Reddit. Moreover, 
the ORIS method starts gaining significant machine performance improvement with budget utilization of 150 documents, and visible human performance gain with the budget utilization of as low as 50. It shows that 
the labelling errors
eventually hamper the machine's performance in the long run. Finally, the inference speed of the proposed ORIS method is close to real-time, outperforming traditional baselines. 
%
Future research could test ORIS for other labeling tasks and 
study the effect of its different hyperparameters in building a robust and inclusive online active learning-based HITL-ML system for streaming analytics. 

\textit{\textbf{Acknowledgment.}} Authors thank the ORIEI grant 215135 at George Mason University to partially support this research. 


\bibliographystyle{IEEEtran}
\bibliography{references}

\end{document}